\documentclass{sigchi}
\usepackage{siunitx}
\usepackage{float}
\usepackage{subfig}
\usepackage[show]{chato-notes} % Change the ``show'' to ``hide'' for submission to remove all \todo and \note's.
\usepackage{amsmath}
\usepackage[titlenumbered,ruled]{algorithm2e}
\usepackage{algpseudocode}
\usepackage{pifont}
\usepackage{times}
\usepackage{booktabs}
\usepackage{multirow}
\usepackage{balance}  % to better equalize the last page
\usepackage{graphics} % for EPS, load graphicx instead 
\usepackage{txfonts}
\usepackage{mathptmx}
\usepackage[pdftex]{hyperref}
\usepackage{color}
\usepackage{booktabs}
\usepackage{textcomp}
\usepackage{microtype} % Improved Tracking and Kerning
\usepackage{ccicons}  % Cite your images correctly!
\usepackage[utf8]{inputenc}
\usepackage[brazilian]{babel}
\usepackage{graphicx}

\newcommand{\green}[1]{\textcolor{green}{#1}}
\newcommand{\red}[1]{\textcolor{red}{#1}}
\newcommand{\blue}[1]{\textcolor{blue}{#1}}

\newcolumntype{P}[1]{>{\centering\arraybackslash}p{#1}}
\newcolumntype{M}[1]{>{\centering\arraybackslash}m{#1}}

\def\plaintitle{}

\def\emptyauthor{}
\def\plainkeywords{}

\makeatletter
\def\url@leostyle{%
  \@ifundefined{selectfont}{
    \def\UrlFont{\sf}
  }{
    \def\UrlFont{\small\bf\ttfamily}
  }}
\makeatother
\def\pprw{8.5in}
\def\pprh{11in}

\definecolor{linkColor}{RGB}{6,125,233}
\hypersetup{%
  pdftitle={\plaintitle},
  pdfauthor={\emptyauthor},
  pdfkeywords={\plainkeywords},
  bookmarksnumbered,
  pdfstartview={FitH},
  colorlinks,
  citecolor=black,
  filecolor=black,
  linkcolor=black,
  urlcolor=linkColor,
  breaklinks=true,
}

\begin{document}

\title{SentiBench - a benchmark comparison of state-of-the-practice sentiment analysis methods}

%\titlenote{\textcolor{red}{\textbf{This is a pre-print of a paper accepted to appear at ASONAM'16.}}}}

\numberofauthors{5}
\author{%
\alignauthor
Filipe N Ribeiro\\
       \affaddr{Federal University of Minas Gerais}\\
       \affaddr{Belo Horizonte, Brazil}\\
       \email{filiperibeiro@dcc.ufmg.br}
\alignauthor
Matheus Araújo\\
       \affaddr{Federal University of Minas Gerais}\\
       \affaddr{Belo Horizonte, Brazil}\\
       \email{fabricio@dcc.ufmg.br}
\alignauthor
Pollyanna Gon\c{c}alves\\
       \affaddr{Federal University of Minas Gerais}\\
       \affaddr{Belo Horizonte, Brazil}\\
       \email{fabricio@dcc.ufmg.br}              
\alignauthor
Marcos Andr{\'e} Gon{\c c}alves\\
       \affaddr{Federal University of Minas Gerais}\\
       \affaddr{Belo Horizonte, Brazil}\\
       \email{fabricio@dcc.ufmg.br}  
\alignauthor
Fabrício Benevenuto\\
       \affaddr{Federal University of Minas Gerais}\\
       \affaddr{Belo Horizonte, Brazil}\\
       \email{fabricio@dcc.ufmg.br}       
}

\maketitle

\begin{abstract}
In the last few years thousands of scientific papers have investigated sentiment analysis, several startups that measure opinions on real data have emerged and a number of innovative products related to this theme have been developed. There are multiple methods for measuring sentiments, including lexical-based and supervised machine learning methods. Despite the vast interest on the theme and wide popularity of some methods, it is unclear which one is better for identifying the polarity (i.e., positive or negative) of a message. Accordingly, there is a strong need to conduct a thorough apple-to-apple comparison of sentiment analysis methods, \textit{as they are used in practice}, across multiple datasets originated from different data sources. Such a comparison is key for understanding the potential limitations, advantages, and disadvantages of popular methods. This article aims at filling this gap by presenting a benchmark comparison of twenty-four popular sentiment analysis methods (which we call the state-of-the-practice methods). Our evaluation is based on a benchmark of eighteen labeled datasets, covering messages posted on social networks, movie and product reviews, as well as opinions and comments in news articles.  Our results highlight  the extent to which the prediction performance of these methods varies considerably across datasets. Aiming at boosting the development of this research area, we open the methods' codes and datasets used in this article, deploying them in a benchmark system, which provides an open API for accessing and comparing sentence-level sentiment analysis methods.
\end{abstract}

\section*{Introduction}

\begin{figure}[h!]
\centering
\includegraphics[width=8.5cm,height=\textheight,keepaspectratio]{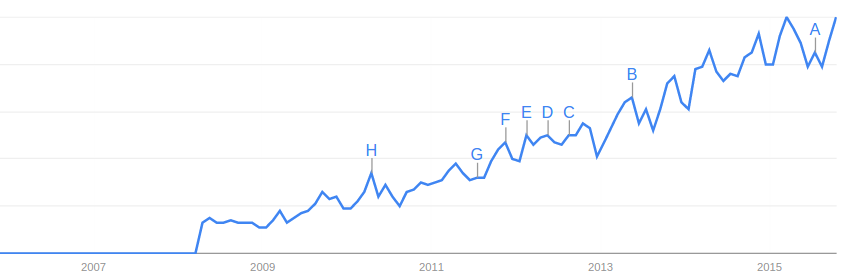}
\caption{Searches on Google for the Query: ``Sentiment Analysis''.}
\label{fig:googletrend}
\end{figure}

Sentiment analysis has become an extremely popular tool, applied in several analytical domains, especially on the Web and social media. To illustrate the growth of interest in the field, Figure~\ref{fig:googletrend} shows the steady growth on  the number of searches on the topic, according to Google Trends\footnote{\url{https://www.google.com/trends/explore\#q=sentiment\%20analysis}}, mainly after the popularization of online social networks (OSNs). More than 7,000 articles have been written about sentiment analysis and various startups are developing tools and strategies to extract sentiments from text~\cite{Feldman:2013:TAS:2436256.2436274}.

The number of possible applications of such a technique is also considerable. Many of them are focused on monitoring the reputation or opinion of a company or a brand with the analysis of reviews of consumer products or services~\cite{Hu:2004:MSC:1014052.1014073}. Sentiment analysis can also provide analytical perspectives  for financial investors who want to discover and respond to market opinions~\cite{conf/epia/OliveiraCA13,DBLP:journals/corr/abs-1010-3003}. Another important set of applications is in politics, where marketing campaigns are interested in tracking sentiments expressed by voters associated with candidates~\cite{Tumasjan}.

%Other examples include: (i) the use of  LIWC \cite{liwc} by Facebook researchers to assess the polarity of user posts to detect if emotional contagion occurs in social networks by means of the manipulation of  news feeds~\cite{Kramer2014}; (ii) the use of  LIWC's polarity prediction over tweets to show mood deterioration across the day \cite{golder-science2011}; (iii) the use of  Sentistrength's~\cite{sentistrength1} polarity detection to build a tool, called Magnet News,  to allow newspapers' readers to choose whether they want to read good or bad news \cite{reis2014icwsm}; (iv) the use of  OpinionFinder \cite{OpinionFinder} to improve the rank of a blog search engine aiming to identify blogs that express an opinion about a target entity \cite{He:2008:ROB:1390334.1390473} and as a tool to define overall blog posts polarities \cite{Chenlo:2011:EEP:2063576.2063634}.

%T

%Several methods exploit a variant of the problem, namely polarity detection --  the degree to which a given message express a  positive or negative opinion about a topic~\cite{Pang+Lee:08b}. In practice, it is usually used to measure the sentiment of small sets of sentences in which the topic is known \textit{a priori}. Polarity detection has numerous applications, for instance, in real time systems  trying to produce social network analytics on product launches~\cite{hannak-2012-weather}. This is the focus of our work.
Due to the enormous interest and applicability, there has been a corresponding increase in the number of proposed sentiment analysis methods in the last years.
The proposed methods rely on many different techniques from different computer science fields.
Some of them employ machine learning methods that often rely on supervised classification approaches, requiring labeled data to train classifiers \cite{Pang:2002:TUS:1118693.1118704}.
Others are lexical-based methods that make use of predefined lists of words, in which each word is associated with a specific sentiment. The lexical methods vary according to the context in which they were created. For instance, LIWC \cite{liwc} was originally proposed to analyze sentiment patterns in formally written English texts, whereas PANAS-t \cite{polly@panast} and POMS-ex \cite{Bollen} were proposed as psychometric scales adapted to the Web context.

Overall, the above techniques are acceptable by the research community and it is common to see concurrent important papers, sometimes published in the same computer science conference, using completely different methods.
For example, the famous Facebook experiment~\cite{Kramer2014} which manipulated users feeds to study emotional contagion, used LIWC~\cite{liwc}. Concurrently,
Reis \textit{et al.} used SentiStrength~\cite{sentistrength1} to measure the negativeness or positiveness of online news headlines~\cite{reis2014icwsm,reis2015@icwsm}, whereas
Tamersoy~\cite{Tamersoy:2015} explored VADER's lexicon~\cite{hutto2014vader} to study patterns of smoking and drinking abstinence in social media.

As the state-of-the-art has not been clearly established, researchers tend to accept any popular method as a valid  methodology to measure sentiments.
However, little is known about the relative performance of the several  existing sentiment analysis methods. In fact, most of the newly proposed methods are rarely compared with all other pre-existing ones using a large number of existing datasets. This is a very unusual situation from a scientific perspective, in which benchmark comparisons are the  rule. In fact, most  applications and experiments reported in the literature make use of previously developed methods exactly how they were released with no changes and adaptations and with none or almost none parameter setting. In other words, the methods have been used as a black-box, without a deeper investigation on their suitability to a particular context or application.

To sum up, existing methods have been widely deployed for developing applications without a deeper understanding regarding their applicability in different contexts or their advantages, disadvantages, and limitations in comparison with each another. Thus, there is a strong need to conduct a thorough apple-to-apple comparison of sentiment analysis methods, \textit{as they are used in practice}, across multiple datasets originated from different data sources.

%\green{Magnet News authors have performed previously experiments to choose the best method to use in that context but we notice that, in general, authors does not investigate which one can achieve better results.}

This  \textit{state-of-the-practice} situation is what we propose to investigate in this article. We do this by providing a thorough benchmark comparison of \textit{twenty-four} \textit{state-of-the-practice} methods using  \textit{eighteen} labeled datasets. In particular, given the recent popularity of online social networks and of short texts on the Web, many methods are focused in detecting sentiments at the sentence-level, usually used to measure the sentiment of small sets of sentences in which the topic is known \textit{a priori}. We focus on such context -- thus, our datasets cover messages posted on social networks, movie and product reviews, and opinions and comments in news articles, Ted talks, and blogs. We  survey an extensive literature on sentiment analysis to identify existing sentence-level methods covering several different techniques. We contacted authors asking for their codes when available or we implemented existing methods when they were unavailable but could be reproduced based on their descriptions in  the original published paper.  We should emphasize that our work focus on off-the-shelf methods as they are used in practice. This excludes most of the supervised methods which require labeled sets for training, as these are usually not available for practitioners. Moreover, most of the supervised solutions do not share the source code or a trained model to be used with no supervision.

%As a result, we provide a benchmark comparison of \textbf{twenty-two} sentence-level sentiment analysis methods.

%There are different problems approached within the field of sentiment analysis, including sentence-level, document-level, and aspect-level sentiment analysis.

%This has numerous applications, for instance, for real time systems  trying to produce social network analytics on product launches~\cite{hannak-2012-weather}. This is the focus of our work.

%Some of them employ machine learning methods that often rely on supervised classification approaches, requiring labeled data to train classifiers~\cite{Pang:2002:TUS:1118693.1118704} and even emerging learning techniques like deep-learning~\cite{Socher-etal:2013}. Others are lexical-based methods that make use of predefined lists of words, in which each word is associated with a specific sentiment. The lexical methods vary according to the context in which they were created. For instance, LIWC~\cite{liwc} was originally proposed to analyze sentiment patterns in formally written English texts, whereas PANAS-t~\cite{polly@panast} and POMS-ex~\cite{Bollen} were proposed as psychometric scales adapted to the Web context.

%In practice, these methods have been proposed independently and comparisons among them are limited, provided partially, and with limited datasets.

%\blue{Introduce main goal: Black box test of 21 methods and comparisons. use next phrases} \yellow{More important, m}

%To approach this problem, we created a benchmark that consists of

Our experimental results unveil a number of important findings. First, we show that there is no single method that always achieves the  best prediction performance for all different datasets, a result consistent with the ``there is no free lunch theorem''~\cite{Wolpert97nofree}. We also show that existing methods vary widely regarding their agreement, even across similar datasets. This suggests that the same content could be interpreted very differently depending on the choice of a sentiment method. We noted that  most methods are more accurate in correctly classifying positive
than negative text, suggesting that current approaches tend to be  biased in their analysis towards positivity.
Finally, we quantify the relative prediction performance of existing efforts in the field across different types of datasets, identifying those with higher prediction performance across different datasets.

% and that can correctly identify positive, neutral, and negative messages accurately

Based on these observations, our final contribution consists on releasing our gold standard dataset and the codes of the compared methods\footnote{Except for paid methods}. We also created a Web system through which we allow other researchers to easily use our data and codes to compare results with the existing methods \footnote{http://www.ifeel.dcc.ufmg.br}. More importantly, by using  our system one could easily test which method would be the most suitable to a particular dataset and/or application.  We hope that our tool will not only help
researchers and practitioners for accessing and comparing a wide range of sentiment analysis techniques, but can also help  towards the development of this research field as a whole.

The remainder of this paper is organized as follows. In Section 2, we briefly describe related efforts. Then, in Section 3 we describe the sentiment analysis methods we compare. Section 4 presents the gold standard data used for comparison. Section 5 summarizes our results and findings.  Finally, Section 6 concludes the article and discusses directions for future work.

\section*{Background and Related Work}

Next we discuss important definitions and justify the focus of our benchmark comparison.
We also briefly survey existing related efforts that compare sentiment analysis methods.

\subsection*{Focus on Sentence-Level Sentiment Analysis}
Since sentiment analysis can be applied to different tasks, we restrict our focus on comparing those efforts related to detect the polarity (i.e. positivity or negativity) of a given short text (i.e. sentence-level). Polarity detection is a common function across all sentiment methods considered in our work, providing valuable information to a number of different applications, specially those that explore short messages that are commonly available in social media~\cite{Feldman:2013:TAS:2436256.2436274}.

%Particularly, we consider those methods that can be performed "as-is" or with small adaptations, as we describe in Section 3.

Sentence-level sentiment analysis can be performed with supervision (i.e. requiring labeled training data) or not.
An advantage of supervised  methods is their ability to adapt and create trained models for specific purposes and contexts.
A drawback is the need of labeled data, which might be highly costly, or even prohibitive, for some tasks. On the other hand, the lexical-based methods make use of a pre-defined list of words, where each word is associated with a specific sentiment. The lexical methods vary according to the context in which they were created. For instance, LIWC~\cite{liwc} was originally proposed to analyze sentiment patterns in English texts, whereas PANAS-t~\cite{polly@panast} and POMS-ex~\cite{Bollen} are psychometric scales adapted to the Web context. Although lexical-based methods do not rely on labeled data, it is hard to create a unique lexical-based dictionary to be used for all different contexts.

We focus our effort on evaluating unsupervised efforts as they can be easily deployed in Web services and applications without the need of human labeling or any other type of manual intervention. As described in Section 3, some of the methods we consider have used machine learning to build lexicon dictionaries or even to build models and tune specific parameters. We incorporate those methods in our study, since they have been released as black-box tools that can be used in an unsupervised manner.

\subsection*{Existing Efforts on Comparison of Methods}

Despite the large number of existing methods, only a limited number of them have performed a comparison among sentiment analysis methods, usually with restricted datasets.
Overall, lexical methods and machine learning approaches have been evolving in parallel in the last years, and it comes as no surprise that studies have started to
compare their performance on specific datasets and use one or another strategy as baseline for comparison.  A recent survey summarizes several of these efforts~\cite{Tsytsarau:2012:SMS:2198180.2198208} and conclude that a systematic comparative study that implements and evaluates all relevant algorithms under the same framework is still missing in the literature. As new methods emerge and compare themselves only against one, at most two other methods, using different evaluation datasets and experimental methodologies, it is hard to conclude if a single method triumphs over the remaining ones, or even in specific scenarios.  To the best of our knowledge, our effort is the first of kind to create a benchmark that provides such thorough comparison.

An important  effort worth mentioning consists of an annual workshop -- The International Workshop on Semantic Evaluation (SemEval). It consists of a series of exercises grouped in tracks,  including sentiment analysis, text similarity, among others, that put several together competitors against each other. Some new methods such as Umigon \cite{levallois:2013:SemEval-2013} have been proposed after obtaining good results on some of these tracks. Although, SemEval has been playing an important role for identifying relevant methods, it requires authors  to register for the challenge and many popular methods have not been evaluated in these exercises. Additionally, SemEval labeled datasets are usually focused on one specific type of data, such as  tweets, and do not represent a wide range of social media data.
In our evaluation effort, we consider one dataset from SemEval 2013 and two methods that participated in the competition in that same year.

Ahmadi et al. \cite{ABBASI14.483} performed a comparision of  twitter-based sentiment analysis tools. They selected twenty tools and tested them across five twitter datasets. This benchmark is the work that most approximate from ours, but it is different in some meaningful aspects. Firstly, we embraced distinct contexts such as reviews, comments and social networks aiming at providing a broader evaluation of the tools. 
%We recognize that Twitter is one of the most important and promising context to employ sentiment analysis but these techniques have also been applied in other situations.
Secondly, the methods they selected included supervised and unsupervised approaches which, in our view, could be unfair for the unsupervised ones. Although the results have been presented separately, the supervised methods, as mentioned by authors, required extensive parameter tuning and validation in a training environment. Therefore, supervised approaches tend to adapt to the context they were applied to. As previously highlighted, our focus is on off-the-shelf  tools as they have been extensively and recently used. Many researchers and practitioners have also used supervised approaches but this is out of scope of our work. Finally, most of the unsupervised methods selected in the Twitter Benchmark are paid tools, except from  two of them, both of which were developed as a result of published academic research. Oppositely we made an extensive bibliography review to include relevant academic outcomes without excluding the most used commercial options.

Finally, in a previous effort~\cite{Pollyanna@COSN13}, we compared eight sentence-level sentiment analysis methods, based on one public dataset used to evaluate SentiStrength~\cite{sentistrength1}. This article  largely extends our previous work by comparing a much larger set of methods across many different datasets, providing a much deeper benchmark evaluation of current  popular sentiment analysis methods. The methods used in this paper were also incorporated as part of an existing system, namely iFeel~\cite{araujo2016@icwsm}.

\section*{Sentiment Analysis Methods}

This section provides a brief description of the twenty-four sentence-level sentiment analysis methods investigated in this article. Our effort to identify important sentence-level sentiment analysis methods consisted of systematically search for them in the main conferences in the field and then checking for papers that cited them as well as their own references. Some of the methods are available for download on the Web; others were kindly shared by their authors under request; and a small part of them were implemented by us based on their descriptions in the original paper. This usually happened when  authors shared only the lexical dictionaries they created, letting the implementation of the method that use the lexical resource to ourselves.

%Finally, these methods assume that the existing entities on the text are known \textit{a priori} (e.g. a corpus of tweets related to the launch of a new brand, or comments on specific aspects of a smartphone in an online store).

Table~\ref{tab:methodsdescription} and Table~\ref{tab:methodscomparison} present an overview of these methods, providing a description of each method as well as the techniques they employ (L for Lexicon Dictionary and ML for Machine Learning), their outputs (e.g. -1,0,1, meaning negative, neutral, and positive, respectively), the datasets they used to validate, the baseline methods used for comparison and finally lexicon details, as well as  the Lexicon size column describing the number of terms contained in the method's lexicon. The methods are organized in chronological order to allow a better overview of the existing efforts over the years. We can note that the methods generate different outputs formats. We colored in blue the positive outputs, in black the neutral ones, and in red those that are negative. Note that we included LIWC and LIWC15 entries in Table~\ref{tab:methodscomparison}, which represents the former version, launched in 2007, and the latest version, from 2015, respectively. We considered both versions because the first one was extensively used in the literature. This also allows to compare the improvements between both versions.

\begin{table*}[htpb]
\scriptsize
  \centering
  \caption{Overview of the sentence-level methods available in the literature.}
\begin{tabular}{ | M{2.7cm} | M{11.6cm} | M{0.18cm} | M{0.13cm}|}
\hline
	\textbf{Name} & \textbf{Description} & \textbf{L} & \textbf{ML}  \\ \hline
	Emoticons~\cite{Pollyanna@COSN13} & Messages containing positive/negative emoticons are positive/negative. Messages without emoticons are not classified. & \checkmark &  \\ \hline	
	Opinion Lexicon \cite{Hu:2004:MSC:1014052.1014073} & Focus on Product Reviews. Builds a Lexicon to predict polarity of product features phrases that are summarized to provide an overall score to that product feature. & \checkmark &  \\ \hline	
	Opinion Finder (MPQA) \cite{OpinionFinder} \cite{wilson-wiebe-hoffmann:2005:HLTEMNLP} & Performs subjectivity analysis trough a framework with lexical analysis former and a machine learning approach latter.& \checkmark &  \checkmark \\ \hline
	
	\if 0
	Happiness Index \cite{DoddsP2009Measuring} & Quantifies happiness levels for large-scale texts as lyrics and blogs. It uses ANEW  words \cite{citeulike:3519108} to rank the documents.& \checkmark &   \\ \hline
		\fi
		
	SentiWordNet \cite{sentiwordnet} \cite{sentiwordnet3} & Construction of a lexical resource for Opinion Mining based on WordNet~\cite{wordnet}. The authors grouped adjectives, nouns, etc in synonym sets (synsets) and associated three polarity scores (positive, negative and neutral) for each one. & \checkmark &  \checkmark\\ \hline
	LIWC \cite{liwc} & An acronym for Linguistic Inquiry and Word Count, LIWC is a text analysis paid tool to evaluate emotional, cognitive, and structural components of a given text. It uses a dictionary with words classified into categories (anxiety, health, leisure, etc). An updated version was launched in 2015. & \checkmark &  \\ \hline
	Sentiment140 \cite{Go2009} & Sentiment140 (previously known as "Twitter Sentiment") was proposed as an ensemble of three classifiers (Naive Bayes, Maximum Entropy, and SVM) built with a huge amount of tweets containing emoticons collected by the authors. It has been improved and transformed into a paid tool.  \if 0 In order to avoid providing high weights on the emoticons they were stripped off from sentences in the training step. Features used were unigrams bigrams, and parts of speech \fi  &  & \checkmark \\ \hline	
	SenticNet \cite{senticnet3} &  Uses dimensionality reduction to infer the polarity of common sense concepts and hence provide a resource for mining opinions from text at a semantic, rather than just syntactic level. & \checkmark &  \\ \hline				
	AFINN \cite{nielsen2011new} - A new ANEW  & Builds a Twitter based sentiment Lexicon including Internet slangs and obscene words. AFINN can be considered as an expansion of ANEW \cite{citeulike:3519108}, a  dictionary created to provides emotional ratings for English words. ANEW dictionary rates words in terms of pleasure, arousal and dominance. & \checkmark &   \\ \hline
	SO-CAL \cite{Taboada:2011:LMS:2000517.2000518} & Creates a new Lexicon with unigrams (verbs, adverbs, nouns and adjectives) and multi-grams (phrasal verbs and intensifiers) hand ranked with scale +5 (strongly positive) to -5 (strongly negative). Authors also included part of speech processing, negation and intensifiers.  & \checkmark &\\ \hline	
	Emoticons DS (Distant Supervision)\cite{hannak-2012-weather} & Creates a scored lexicon based on a large dataset of tweets. Its based on the frequency each lexicon occurs with positive or negative emotions.  & \checkmark &  \\ \hline	
	NRC Hashtag \cite{mohammad:2012:STARSEM-SEMEVAL} & Builds a lexicon dictionary using a Distant Supervised Approach. In a nutshell it uses known hashtags (i.e \#joy, \#happy etc) to ``classify'' the tweet. Afterwards, it verifies frequency each specific n-gram occurs in a emotion and calculates its Strong of Associaton with that emotion. & \checkmark &  \\ \hline
	Pattern.en \cite{de2012pattern} & Python Programming Package (toolkit)  to deal with NLP,  Web Mining and Sentiment Analysis. Sentiment analysis is provided through averaging scores from adjectives in the sentence according to a bundle lexicon of adjective. & \checkmark &   \\ \hline
	SASA \cite{sasa} & Detects public sentiments on Twitter during the 2012 U.S. presidential election. It is based on the statistical model obtained from the classifier Na\"{i}ve Bayes on unigram features. It also explores emoticons and exclamations. &  &  \checkmark\\ \hline		
	PANAS-t \cite{polly@panast} & Detects mood fluctuations of users on Twitter. The method consists of an adapted version (PANAS) {Positive Affect Negative Affect Scale} \cite{Watson}, well-known method in psychology with a large set of words, each of them associated with one from eleven moods such as surprise, fear, guilt, etc \if 0(surprise, fear, joviality, assurance, serenity, sadness, guilt, hostility, shyness, fatigue, and attentiveness)\fi. & \checkmark &  \\ \hline
	EmoLex \cite{journals/ci/MohammadT13} & Builds a general sentiment Lexicon crowdsourcing supported. Each entry lists the association of a token with 8 basic sentiments: joy, sadness, anger, etc defined by~\cite{citeulike:8791184}. Proposed Lexicon includes unigrams and bigrams from Macquarie Thesaurus and also words from GI and Wordnet. & \checkmark &   \\ \hline	
	USent \cite{Pappas_CICLING_2013} & Infer additional reviews user ratings by performing sentiment analysis (SA) of user comments and integrating its output in a nearest neighbor (NN) model that provides multimedia recommendations over TED Talks. & \checkmark &  \checkmark\\ \hline
	Sentiment140 Lexicon \cite{MohammadKZ2013} & A lexicon dictionary based on the same dataset used to train the Sentiment140 Method. The lexicon was built in a similar way to \cite{mohammad:2012:STARSEM-SEMEVAL} but authors used the occurrency of emoticons to classify the tweet as positive or negative. Then, the n-gram score was calculated based on the frequency of occurrence in each class of tweets. & \checkmark &  \\ \hline
	SentiStrength \cite{sentistrength1} & Builds a lexicon dictionary annotated by humans and improved with the use of Machine Learning. & \checkmark &  \checkmark \\ \hline
	Stanford Recursive Deep Model \cite{Socher-etal:2013} & Proposes a model called Recursive Neural Tensor Network (RNTN) that processes all sentences dealing with their structures and compute the interactions between them. This approach is interesting since RNTN take into account the order of words in a sentence, which is ignored in most of methods.& \checkmark & \checkmark \\ \hline
	Umigon \cite{levallois:2013:SemEval-2013} & Disambiguates tweets using lexicon with heuristics to detect negations plus elongated words and hashtags evaluation. & \checkmark &   \\ \hline
	ANEW\_SUB \cite{anew_wkb} & Another extension of the ANEW dictionary \cite{citeulike:3519108} including the most common words from the SubtlexUS corpus \cite{subtlex}. SubtlexUS was an effort to propose a different manner to calculate word frequencies considering film and TV subtitles.& \checkmark &  \\ \hline
	VADER \cite{hutto2014vader} & It is a human-validated sentiment analysis method developed for twitter and social media contexts. VADER was created from a generalizable, valence-based, human-curated gold standard sentiment lexicon. & \checkmark &  \\ \hline
	Semantria \cite{semantria2015} & It is a paid tool that employs multi-level analysis of sentences. Basically it has four levels: part of speech, assignment of previous scores from dictionaries, application of intensifiers and finally machine learning techniques to delivery a final weight to the sentence. & \checkmark & \checkmark \\ \hline	
    \end{tabular}
  \label{tab:methodsdescription}
\end{table*}

\begin{table*}[htpb]
\scriptsize
  \centering
  \caption{Overview of the sentence-level methods available in the literature.}
\begin{tabular}{ | M{1.5cm} | M{4.5cm}| M{2.4cm} | M{5.0cm} | M{1.1cm}|}
\hline
	\textbf{Name} & \textbf{Output} & \textbf{Validation} & \textbf{Compared To} & \textbf{Lexicon size}\\ \hline
	Emoticons  & \textbf{\red{-1}, \blue{1}} & - & - & 79 \\ \hline	
	Opinion Lexicon & Provides polarities for lexicons & Product Reviews from Amazon and CNet & - & 6,787  \\ \hline	
	Opinion Finder (MPQA)  & \textbf{\red{Negative}, Objective, \blue{Positive}} & MPQA \cite{Wiebeetal05} & Compared to itself in different versions & 20,611 \\ \hline
\if 0
	13,915
	Happiness Index & \textbf{\red{1}, \red{2}, \red{3}, \red{4}, 5, \blue{6}, \blue{7}, \blue{8}, \blue{9}} & Lyrics, Blogs,STU Messages \footnote{State of Union Messages - http://www.presidency.ucsb.edu/}, British National Corpus \footnote{http://www.natcorp.ox.ac.uk/},  & - & 1,034 \\ \hline
\fi 	
	SentiWordNet & Provides positive, negative and objective scores for each word (0.0 to 1.0 ) \if 0 \textbf{\red{-1}, 0, \blue{1}} \fi & - & General Inquirer (GI)\cite{Stone66} & 117,658\\ \hline
	 Sentiment140 & \textbf{\red{0}, 2, \blue{4}} & Their own datasets - 359 tweets (Tweets\_STF, presented at Table \ref{tab:labeled_dataset}) & Naive Bayes, Maximum Entropy, and SVM classifiers as described in \cite{Pang:2002:TUS:1118693.1118704} & - \\ \hline			
	LIWC07 & \textbf{\red{negEmo}, \blue{posEmo}} & - & Their previous dictionary (2001) & ~ 4,500\\ \hline
	SenticNet &  \textbf{\red{Negative}, \blue{Positive}} & Patient Opinions (Unavailable) & SentiStrength \cite{sentistrength1} & 15,000\\ \hline		
	AFINN & \if 0 \textbf{\red{[-5..)} ,-1..1,  \blue{(..5]}}\textbf{\red{-1}, 0, \blue{1}}\fi Provides polarity score for lexicons (-5 to 5). & Twiter \cite{Biever2010} & OpinonFinder \cite{OpinionFinder}, ANEW \cite{citeulike:3519108}, GI \cite{Stone66} and Sentistrength \cite{sentistrength1} & 2,477 \\ \hline
SO-CAL &  \textbf{\red{[\textless 0)}, 0, \blue{(\textgreater 0]}} & Epinion \cite{Taboada06a}, MPQA\cite{Wiebeetal05}, Myspace\cite{sentistrength1},  &  MPQA\cite{Wiebeetal05}, GI\cite{Stone66}, SentiWordNet \cite{sentiwordnet},"Maryland" Dict \cite{Mohammad-maryland2009}, Google Generated Dict \cite{Taboada:06a} & 9,928 \\ \hline	
	Emoticons DS (Distant Supervision)& Provides polarity score for lexicons  & Validation with unlabeled twitter data \cite{cha_icwsm10} & -  & 1,162,894 \\ \hline	
	NRC Hashtag & Provides polarities for lexicons & Twitter (SemEval-2007 Affective Text Corpus)~\cite{Strapparava:2007} & WordNet Affect~\cite{Strapparava:2007} & 679,468 \\ \hline
	Pattern.en &\textbf{ Objective, \red{[ \textless 0.1}, \blue{$\geq$0.1]}} & Product reviews, but the source was not specified & - & 2,973\\ \hline
	SASA \cite{sasa}& \textbf{\red{Negative}, Neutral, Unsure, \blue{Positive}} & ``Political'' Tweets labeled by ``turkers'' (AMT) (unavailable) & - & - \if 0 21,012 \fi \\ \hline		
	PANAS-t &  Provides association for each word with eleven moods (joviality, attentiveness, fear, etc ) \if 0 \textbf{\blue{joviality, assurance, serenity, and surprise} ,attentiveness, \red{fear, sadness, guilt, hostility, shyness, and fatigue}}\fi & Validation with unlabeled twitter data \cite{cha_icwsm10}  & - & 50 \\ \hline
	EmoLex & \if 0 \textbf{\red{-1}, 0, \blue{1}} \fi Provides polarities for lexicons & - & Compared with existing gold standard data but it was not specified & 141,820\\ \hline	
	USent & \textbf{\red{neg}, neu, \blue{pos}} & Their own dataset - Ted Talks & Comparison with other multimedia recommendation approaches& MPQA (8,226) / Their own (9,176)\\ \hline
	Sentiment140 Lexicon & Provides polarity scores for lexicon \if 0 \textbf{\red{-1},0,\blue{1}} \fi & Twitter and SMS from Semeval 2013, task 2 \cite{semeval-2013task2} &  Other Semeval 2013, task 2 approaches &  1,220,176\\ \hline
	SentiStrength & \textbf{\red{-1},0,\blue{1}} & Their own datasets - Twitter, Youtube, Digg, Myspace, BBC Forums and Runners World & The best of nine Machine Learning techniques for each test& 2,698
 \\ \hline
	Stanford Recursive Deep Model & \textbf{\red{very negative}, \red{negative}, neutral, \blue{positive}, \blue{very positive}} & Movie Reviews \cite{Pang04asentimental} & Na\"{i}ve Bayes and SVM with bag of words features and bag of bigram features & 227,009\\ \hline
	Umigon & \textbf{\red{Negative}, Neutral, \blue{Positive}} & Twitter and SMS from Semeval 2013, task 2 \cite{semeval-2013task2} & \cite{MohammadKZ2013} & 1,053\\ \hline
	 ANEW\_SUB &  Provides ratings for words in terms of Valence, Arousal and Dominance. Results can also be grouped by gender, age and education & - & Compared to similar works, including cross-language studies, by means of correlations between emotional dimensions & 13,915 \\ \hline	
	VADER & \textbf{\red{[\textless -0,05)}, (-0,05..0,05), \blue{(\textgreater 0,05]}} & Their own datasets - Twitter, Movie Reviews, Technical Product Reviews, NYT User's Opinions & (GI)\cite{Stone66}, LIWC, \cite{liwc}, SentiWordNet \cite{sentiwordnet}, ANEW \cite{citeulike:3519108}, SenticNet \cite{FSS102216} and some Machine Learning Approaches & 7,517\\ \hline
	LIWC15 & \textbf{\red{negEmo}, \blue{posEmo}} & - & Their previous dictionary (2007) & ~ 6,400\\ \hline	
	Semantria & \textbf{\red{negative}, neutral, \blue{positive}} & not available & not available & not available\\ \hline
    \end{tabular}
  \label{tab:methodscomparison}
\end{table*}

%TEM QUE EXPLICAR QUE LEXICAL DICTIONARIES NAO SAO NECESSARIAMENTE METODOS, MAS SAO MUITO USADOS COMO TAL. Ex. o liwc. TODOS OS LEXICAL RESOURCES FORAM AVALIADOS NA IMPLEMENTACAO DO VADER.

\subsection*{Adapting Lexicons for the Sentence Level Task}

Since we are comparing sentiment analysis methods on a sentence-level basis, we need to work with mechanisms that are able to receive sentences as input and produce polarities as output. Some of the approaches considered in this paper, shown in Table~\ref{tab:methodscomparison}, are complex dictionaries built with great effort. However, a lexicon alone has no natural ability to infer polarity in sentence level tasks.
The purpose of a lexicon goes beyond the detection of polarity of a sentence~\cite{Feldman:2013:TAS:2436256.2436274,liu2012sentiment}, but it can also be used with that purpose~\cite{godbole2007large,kouloumpis2011twitter}.

Several existing sentence-level sentiment analysis methods, like VADER \cite{hutto2014vader} and SO-CAL \cite{Taboada:2011:LMS:2000517.2000518}, combine a lexicon and the processing of the sentence characteristics to determine a sentence polarity. These approaches make use of a series of intensifiers, punctuation transformation, emoticons, and many other heuristics.

Thus, to evaluate each lexicon dictionaries as the basis for a sentence-level sentiment analysis method, we considered the VADER's implementation. In other words, we used VADER's code for determining if a sentence is positive or not considering different lexicons. The reasons for choosing VADER are twofold: (i) the fact it is an open source tool, allowing easy replication of the procedures we performed in our study; and (ii)  VADER's expressive results observed in previous experiments.

%This usually happened when  authors shared only the lexical dictionaries they created, letting the implementation of the method that use the lexical resource to ourselves. lexicon provides prior polarity for a term that can be used to define the polarity of the whole sentence by a system.

%Accordingly, and for comparative purposes, we have implemented two strategies to detect polarity based on each lexicon. The  first one is the simplistic strategy mentioned above based on summing up prior polarities and the second one employed Vader's General Heuristic described below. In spite of the expectation of poorly results, the first approach was developed to be used as baseline for comparison.

VADER's heuristics were proposed based on qualitative analyses of textual properties and  characteristics which affect the perceived sentiment intensity of the text. VADER's author identified five heuristics based on grammatical and syntactical cues to convey changes to sentiment intensity that go beyond the bag-of-words model. The heuristics include treatments for: 1) punctuation (e.g number of `!'s); 2) capitalization (e.g "I HATE YOU" is more intense than "i hate you"); 3) degree modifiers (e.g "The service here is extremely good" is more intense than "The service here is good"); 4) constructive conjunction "but" to shift the polarity; 5) Tri-gram examination to identify negation (e.g "The food here isn’t really all that great."). We choose VADER as a basis for such heuristics as it is one of the most recent methods among those we considered. Moreover, it is becoming widely used, being even implemented as part of the well-known NLTK python library\footnote{http://www.nltk.org/\_modules/nltk/sentiment/vader.html}.

We applied such heuristics to the following lexicons: ANEW\_SUB, AFINN, Emolex, EmoticonsDS, NRC Hashtag, Opinion Lexicon, PANAS-t, Sentiment 140 Lexicon and SentiWordNet. We notice that those strategies drastically improved most of the results of the lexicons for sentence-level sentiment analysis in comparison with a simple baseline approach that averages the occurrence of positive and negative words to classify the polarity of a sentence. The results for the simplest usage of the above lexicons as plain methods are available in the last four Tables in the Additional File 1 of the electronic version of the manuscript. LIWC dictionary was not included in these adaptations due to its very restrictive license, which  does not allow any derivative work based on the original application and lexicon. Table~\ref{tab:methodscomparison}  has also a column (Lexicon size) that describes the number of terms contained in the the proposed dictionary.

\if 0
, except from LIWC. Since LIWC has been extensively used by practitioners and researchers as a tool for sentiment analysis we decided to mantain their original results in the main results' Table.} 
}
\fi
 %contains and column C (changed) indicates some methods we slightly modified to adequate their output formats to the polarity detection task, as described next. FILIPE: ESTA COLUNA NÃO EXISTE MAIS. CREIO QUE NO PASSADO RESOLVEMOS TIRÁ-LA JÁ QUE NA SEÇÃO SEGUINTE DESCREVEMOS AS ADAPTAÇÕES}

\subsection*{Output Adaptations}

It is worth noticing that the output of each method varies drastically depending on the goal it was developed for and the approach it employs. PANAS-t, for instance, associates each word with eleven moods as described in Table~\ref{tab:methodsdescription} and it was designed to track any increase or decrease in sentiments over time. Emolex lexicon provides the association of each word with eight sentiments. The word 'unhappy' for example  is related to anger, disgust, and sadness and it is not related to joy, surprise, etc. SentiWordNet links each word with a synset (i.e. a set of synonyms) characterized by a positive and a negative score, both of them represented with a value between 0 and 1.

The aforementioned lexicons were used as dictionary input to VADER's code. We had to adapt the way the words are processed as follows. For PANAS-t we assumed that joviality, assurance, serenity, and surprise are positive affect. Fear, sadness, guilt, hostility, shyness, and fatigue are negative affect. Attentiveness was considered neutral. In the case of Emolex, we considered two other entries released by the authors. The first one defines the positiviy of a word (0 or 1) and the second characterizes the negativity (0 or 1). For SentiWordNet we calculate an overall score to the word by subtracting the positive value from negative value defined to that word. For example, the positive value for the word faithful is 0.625 while its negative score is 0.0. Then the overall score   score is 0.625. Finally, for ANEW\_SUB we employed only the valence emotion of each word. This metric  ranges from 1 to 9 and indicates the level of pleasantness of a specific word -- we considered the values one to four as negative, five as neutral, and six to nine as positive.

\if 0 
In the case of LIWC, we used the words presented in the positive and negative categories, assigning weights 1 and -1 depending on the category the word belonged to.
\fi

Other lexicons included in our evaluation already provide positive and negative scores such as SentiWordNet or an overall score ranging from a negative to a positive value. After applying VADER's heuristics for each one of these lexicons we get scores in the same way VADER's output (see Table~\ref{tab:methodscomparison}).

\if 0
The aforementioned lexicons were used as dictionary input to Vader's code, we have to adapt the way the words are processed, as follows:
\begin{itemize}
\item For PANAS-t we assumed that joviality, assurance, serenity, and surprise are positive affect, fear, sadness, guilt, hostility, shyness, and fatigue are negative affect, and attentiveness is neutral. If the word is related to a positive, negative or neutral sentiment its score is 1,-1 and 0 respectively.
\item  In the case of Emolex, we used two other entries for each word in the lexicon that relates the word to positive or negative sentiment and has value 0 or 1 for each sentiment. 
\item For SentiWordNet we subtract positive score from negative score and provide the resultant value for Vader's code. 
\item Other lexicons included in our evaluation already provide positive and negative scores like SentiWordNet or an overall score ranging from a negative to a positive value.
\end{itemize} 
 After applying Vader's heuristics for each one of these lexicons we get -1 for negative sentences, 0 for neutral and 1 for positive in the same way Vader's output (see Table~\ref{tab:methodscomparison}). 

\fi

Other methods also required some output handling. The available implementation of OpinionFinder~\footnote{http://mpqa.cs.pitt.edu/opinionfinder/}, for instance, generates polarity outputs (-1,0, or 1) for each sentiment clue found in a sentence so that a single sentence can have more than one clue.  We considered the polarity of a single sentence as the sum of the polarities of all the clues. 
%\red{Happiness index uses scores ranging from 1 to 9 indicating the level of happiness in the text -- we considered the values one to four as negative, five as neutral, and six to nine as positive. REMOVER - HAP. INDEX NÃO SERÁ MAIS UTILIZADO COMO MÉTODO} 

The outputs from the remaining methods were easily adapted and converted to positive, negative or neutral. SO-CAL and Pattern.en delivery float numbers greater than a threshold, indicating positive, and lesser than the threshold, indicating negative. LIWC, SenticNet, SASA, USent, SentiStrength, Umigon, VADER and Semantria already provide fixed outputs indicating one of three desired classes while Stanford Recursive Deep Model yields very negative and very positive which in our experiments are handled as negative and positive, respectively.

\subsection*{Paid Softwares}

Seven out of the twenty-four methods evaluated in this work are closed paid softwares: LIWC (2007 and 2015), Semantria, SenticNet 3.0, Sentiment140 and SentiStrength. Although SentiStrength is paid, it has a free of charge academic license. SenticNet's authors kindly processed all datasets with the commercial version and return the polarities for us. For SentiStrenght we used the Java version from May 2013 in a package with all features of the commercial version. For LIWC we acquired the licenses from 2007 (LIWC07) and 2015 (LIWC15) versions.  Finally, for Semantria and Sentiment140 we used a trial account free of charge for a limited number of sentences, which was sufficient to run our experiments.  
   
\subsection*{Methods not included}

Despite our effort to include in our comparison most of the highly cited and important methods we could not  include a few of them for different reasons. Profile of Mood States (POMS-ex)~\cite{Bollen} is not available on the Web or under request and could not be re-implemented based on their descriptions in the original papers. The same situation occurs with the Learning Sentiment-Specific Word Embedding for Twitter Sentiment Classification ~\cite{TangWYZLQ14}. NRC SVM~\cite{MohammadKZ2013} is not available as well, although the lexical resources used by the authors are available and were considered in our evaluation resulting in the methods: NRC Hashtag and Sentiment140. The authors of the Convolutional Neural Network for Modeling Sentences~\cite{KalchbrennerACL2014} and of the Effective Use of Word Order for Text Categorization with Convolutional Neural Networks~\cite{Johnson015} have made their source code available but the first one lacks the train files and the second one requires a GPU to execute. There are a few other methods for sentiment detection proposed in the literature and not considered here. Most of them consists of variations of the techniques used by the above methods, such as WordNet-Affect\cite{Valitutti04wordnet-affect:an} and Happiness Index \cite{DoddsP2009Measuring}. 

\subsection*{Datasets and Comparison Among Methods}

%We are currently contacting some authors of methods to show them our effort and hear feedback, hoping to be able to fix any issue in time for publication.

From Table~\ref{tab:methodscomparison} we can note that the validation strategy, the datasets used, and the comparison with baselines performed by these methods vary greatly, from toy examples to large labeled datasets. PANAS-t and Emoticons DS used manually unlabeled twitter data to validate their methods, by presenting evaluations of events in which some bias towards positivity and negativity would be expected. PANAS-t is tested with unlabeled Twitter data related to Michael Jackson's death and the release of a Harry Potter movie whereas Emoticons DS verified the influence of weather and time on the aggregate sentiment from Twitter.  Lexical dictionaries were validated in very different ways. AFINN\cite{nielsen2011new} compared its Lexicon with other dictionaries. Emoticon Distance Supervised~\cite{hannak-2012-weather} used Pearson Correlation between human labeling and the predicted value. SentiWordNet~\cite{sentiwordnet}  validates the proposed dictionary with comparisons with other dictionaries, but it also used human validation of the proposed lexicon. These efforts attempt to validate the created lexicon, without comparing the lexicon as a sentiment analysis method by itself. VADER~\cite{hutto2014vader} compared results with lexical approaches considering labeled datasets from different social media data. SenticNet \cite{FSS102216} was compared with SentiStrength~\cite{sentistrength1} with a specific dataset related to patient opinions, which could not be made available. Stanford Recursive Deep Model~\cite{Socher-etal:2013} and SentiStrength~\cite{sentistrength1} were both compared with standard machine learning approaches, with their own datasets.
%In the specific case of EmoLex, authors mentioned a comparison with existing gold standard data but it was not directly specified.

This scenario, where every new developed solution compares itself with different solutions using different datasets,
happens because  there is no standard benchmark for evaluating new methods. This problem is exacerbated because many methods have been proposed in different research communities (e.g. NLP, Information Science, Information Retrieval, Machine Learning), exploiting different techniques, with low knowledge about related efforts in other communities.  Next, we describe how we created a large gold standard to properly compare all the considered sentiment analysis methods.

\section*{Gold Standard Data}

A key aspect in evaluating sentiment analysis methods consists of using accurate gold standard labeled datasets. Several existing efforts have generated labeled data produced by experts or  non-experts evaluators. Previous studies suggest that both efforts are valid as non-expert labeling may be  as effective as annotations produced by experts for affect recognition, a very related task~\cite{Snow:2008:CFG:1613715.1613751}. Thus, our effort to build a large and representative gold standard dataset consists of obtaining labeled data from trustful previous efforts that cover a wide range of sources and kinds of data. We also attempt to assess
the ``quality'' of our gold standard in terms of the accuracy of the labeling process.

Table~\ref{tab:labeled_dataset} summarizes the main characteristics of the eighteen  exploited datasets, such as number of messages and the average number of words per message in each dataset.  It also defines a simpler nomenclature that is used in the remainder of this paper.
The table also presents the methodology employed in the classification. Human labeling was implemented in almost all datasets, usually done with the use of non-expert reviewers.
Reviews\_I dataset relies on five stars rates, in which users rate and provide a comment about an entity of interest (e.g. a movie or an establishment).
%Reviews\_I and YELP datasets used these user ratings to determine if a comment is positive or negative.

\begin{table*}[htpb]
  \centering
	\scriptsize
  \caption{Labeled datasets.}
    \begin{tabular}{|M{4.2cm}|M{1.9cm}|M{0.4cm}|M{0.4cm}|M{0.4cm}|M{0.4cm}|M{1.3cm}|M{1.1cm}|M{1.8cm}|M{1.2cm}|M{0.5cm}|}
		\hline
    \textbf{Dataset} & \textbf{Nomeclature} & \textbf{     \#} & \textbf{   \#} & \textbf{   \# } & \textbf{   \# } & \textbf{Average \#} & \textbf{Average \#} & \textbf{Annotators} & \textbf{\# of} & \textbf{CK}\\
   &  & \textbf{Msgs}  & \textbf{Pos} & \textbf{Neg}& \textbf{Neu}& \textbf{of phrases} & \textbf{of words} & \textbf{Expertise} & \textbf{Annotators} &  \\ \hline
   
		\textbf{Comments (BBC)} \cite{sentistrength1} & \textbf{Comments\_BBC} &  1,000 & 99 & 653 & 248 & 3.98  & 64.39 & Non Expert & 3 & 0.427 \\
 \hline
		\textbf{Comments (Digg)} \cite{sentistrength1} & \textbf{Comments\_Digg} &  1,077 & 210 & 572 & 295 & 2.50 & 33.97 & Non Expert & 3 & 0.607\\
 \hline
		\textbf{Comments (NYT)} \cite{hutto2014vader}& \textbf{Comments\_NYT} &  5,190 & 2,204 & 2,742 & 244 & 1.01 & 17.76 & AMT & 20 & 0.628\\
\hline			
		\textbf{Comments (TED)} \cite{pappas2013sentiment}  & \textbf{Comments\_TED} & 839 & 318 & 409 & 112 & 1 & 16.95 & Non Expert& 6 & 0.617\\
\hline	
		\textbf{Comments (Youtube)} \cite{sentistrength1} & \textbf{Comments\_YTB} & 3,407 & 1,665 & 767 & 975 & 1.78 & 17.68 & Non Expert& 3 & 0.724\\
\hline
		\textbf{Movie-reviews} \cite{Pang04asentimental}  & \textbf{Reviews\_I} &  10,662 & 5,331 & 5,331 & - & 1.15 & 18.99 & \ User Rating & - & 0.719\\
\hline	
		\textbf{Movie-reviews} \cite{hutto2014vader} & \textbf{Reviews\_II} &  10,605 & 5,242 & 5,326 & 37 & 1.12 & 19.33 & AMT & 20 & 0.555\\
\hline
		\textbf{Myspace posts} \cite{sentistrength1} & \textbf{Myspace} &  1,041 & 702 & 132 & 207 & 2.22 & 21.12 & Non Expert& 3 & 0.647\\
\hline		
		\textbf{Product reviews} \cite{hutto2014vader} & \textbf{Amazon} & 3,708 & 2,128 & 1,482 & 98 & 1.03 & 16.59 & AMT & 20 & 0.822 \\
\hline		
		\textbf{Tweets (Debate)} \cite{diakopoulos2010characterizing} & \textbf{Tweets\_DBT} & 3,238 & 730 & 1249 & 1259 & 1.86 & 14.86 & AMT + Expert & Undef. & 0.419 \\
\hline			
\if 0		
\textbf{Tweets (Irony)} & \textbf{Irony} &  100 & 38 & 43 & 19 & 1,01 & 17,44  & Expert & 3 & - \\
		(Labeled by us) &  &   &  &  & &   &  & & &
		\\ \hline
		\textbf{Tweets (Sarcasm)} & \textbf{Sarcasm} &  100 & 38 & 38 & 24 & 1 & 15,55 & Expert & 3 & -\\
		(Labeled by us) &  &   &  &  & &  &  &  & &
		\\ \hline
\fi 		
		\textbf{Tweets (Random)} \cite{sentistrength1} & \textbf{Tweets\_RND\_I} &  4,242 & 1,340 & 949 & 1953 & 1.77 & 15.81 & Non Expert & 3 & 0.683\\
\hline		
		\textbf{Tweets (Random)} \cite{hutto2014vader} & \textbf{Tweets\_RND\_II} & 4,200 & 2,897 & 1,299 & 4 & 1.87  & 14.10 & AMT & 20 & 0.800\\
\hline		
		\textbf{Tweets (Random)} \cite{SaschaNarr2012}& \textbf{Tweets\_RND\_III} & 3,771 & 739 & 488 & 2,536 & 1.54 & 14.32 & AMT & 3 & 0.824\\
\hline				
		\textbf{Tweets (Random)} \cite{aisopos2014}& \textbf{Tweets\_RND\_IV} & 500 & 139 &119 & 222 & 1.90 & 15.44 & Expert & Undef. &  0.643\\
\hline
		\textbf{Tweets (Specific domains w/ emot.)} \cite{Go2009}& \textbf{Tweets\_STF} & 359 & 182 & 177 & - & 1.0 & 15.1 & Non Expert & Undef. & 1.000\\
\hline	
		\textbf{Tweets (Specific topics)} \cite{sanders2011}& \textbf{Tweets\_SAN} & 3737 & 580 & 654 & 2503 & 1.60 & 15.03 & Expert & 1 & 0.404\\
\hline				
		\textbf{Tweets (Semeval2013 \- Task2)} \cite{semeval-2013task2}& \textbf{Tweets\_Semeval} & 6,087 & 2,223 & 837 & 3027 & 1.86 & 20.05  & AMT & 5 & 0.617\\
\hline		
		\textbf{Runners World forum} \cite{sentistrength1}  & \textbf{RW} &  1,046 & 484 & 221 & 341 & 4.79 & 66.12 & Non Expert& 3 & 0.615\\
\hline		
\if0		\textbf{Yelp Dataset}& \textbf{YLP} &  5,000 & 2,500 & 2,500 & - & 1  & 131.44 & User& - & 94.0\\ 							\cite{Yelp}  &  &   &  &  &  &  &  & Rating & &
		\\ \hline
\fi
		\end{tabular}
		
  \label{tab:labeled_dataset}
\end{table*}

Labeling based on Amazon Mechanical Turk (AMT) was used in seven out of the eighteen datasets, while volunteers and other strategies that involve non-expert evaluators were used in ten datasets. Usually, an agreement strategy (i.e. majority voting) is applied to ensure that, in the end, each sentence has an agreed-upon polarity assigned to it. The number of annotators used to build the datasets is also shown in Table~\ref{tab:labeled_dataset}.

Tweets\_DBT was the unique dataset built with a combination of AMT Labeling with Expert validation~\cite{diakopoulos2010characterizing}. They selected 200 random tweets to be classified by experts and compared with AMT results to ensure accurate ratings. We note that the Tweets\_Semeval dataset was provided as a list of Twitter IDs, due to the Twitter policies related to data sharing. While crawling the respective tweets, a small part of them could not be accessed, as they were deleted. We plan to release all gold standard datasets in a request basis, which is in agreement with Twitter policies.

In order to assess the extent to which these datasets are trustful, we used a strategy similar to the one used by Tweets\_DBT. Our goal was not to redo all the performed human evaluation, but simply inspecting a small sample of them to infer the level of agreement with our own evaluation. We randomly select 1\% of all sentences to be evaluated by experts (two of the authors) as an attempt to assess if these gold standard data are really trustful. It is important to mention that we did not have access to the instructions provided by the authors. We also could not get access to small amount of the raw data in a few datasets, which was discarded. Finally, our manual inspection  unveiled a few sentences in  idioms other than English in a few datasets, such as Tweets\_STA and TED, which were obviously discarded.

% We also attempted to identify the messages that were suspect to be in different languages in the rest of the datasets. Then we manually inspected the suspected ones and removed those that are not in English.

Column CK from Table~\ref{tab:labeled_dataset} exhibits the level of agreement of each dataset in our evaluation by means of Cohen's Kappa, an extensively used metric to calculate inter-anotator agreement.  After a close look in the cases of disagreement with the evaluations in the Gold standard, we realized that other interpretations could be possible for the given text, finding cases of sentences with mixed polarity. Some of them are strongly linked to original context and are very hard to evaluate. Some NYT comments, for instance, are directly related to the news they were inserted to. We can also note that some of the datasets do not contain neutral messages. This might be a characteristic of the data or even a result of how annotators were instructed to label  their pieces of text. Most of the cases of disagreement involve neutral messages. Thus, we considered these cases, as well as the amount of disagreement we had with the gold standard data, reasonable and expected, specially when taking into account that Landis and Koch \cite{Landis77} suggest that Kappa values between 0.4 and 0.6 indicate moderate agreement and values amid 0.60 and 0.8 correspond to substantial agreements.

\section*{Comparison Results}

%\red{
%\begin{itemize}
%  \item no silver bullet
%  \item Metodos que perdem para o random
%  \item Mostra as flutuacoes no twitter
%  \item Metodos melhores positivos, quando negativos.
%  \item tabela media de macro e coverage
%\end{itemize}
%}

Next, we present comparison results for the twenty-four methods considered in this paper based on the eighteen considered gold standard datasets.

\subsection*{Experimental details}

At least three distinct approaches have been proposed to deal with sentiment analysis of sentences. The first of them, applied by OpinionFinder and Pattern.en, for instance, splits this task into two steps: (i) identifying sentences with no sentiment, also named as objective vs. neutral sentences and then (ii) detecting the polarity (positive or negative), only for the subjective sentences. Another common way to detect sentence polarity considers three distinct classes (positive, negative and neutral) in a single task, an approach used by VADER, SO-CAL, USent and others. Finally, some methods like SenticNet and LIWC, classify a sentence as positive or negative only, assuming that only polarized sentences are presented, given the context of a given application. As an example, reviews of products are expected to contain only polarized opinion.

Aiming at providing a more thorough comparison among these distinct approaches, we perform two rounds of tests. In the first we consider the performance of methods to identify 3-class (positive, negative and neutral). The second considers only positive and negative as output and assumes that a previous step of removing the neutral messages needs to be executed firstly. In the 3-class experiments we used only datasets containing a considerable number of neutral messages (which excludes Tweets\_RND\_II, Amazon, and Reviews\_II). Despite being  2-class methods, as highlighted in Table~\ref{tab:methodscomparison}, we decided to include LIWC, Emoticons and SenticNet in the 3-class experiments to present a full set of comparative experiments.  LIWC, Emoticons, and SenticNet cannot define, for some sentences, their positive or negative polarity, considering it as undefined. It occurs due to the absence in the sentence of emoticons (in the case of Emoticons method) or of words belonging to the methods' sentiment lexicon. As neutral (objective) sentences do not contain  sentiments, we assumed, in the case of these 2-class methods, that sentences with undefined polarities are equivalent to neutral sentences.

The 2-class experiments, on the other hand, were  performed with all datasets described in Table~\ref{tab:labeled_dataset} excluding the neutral sentences. We also included all methods in these experiments, even those that produce neutral outputs. As discussed before, when 2-class methods cannot detect the polarity (positive or negative) of a sentences they usually assign it to an undefined polarity. As we know all sentences in the 2-class experiments are positive or negative, we create the coverage metric to determine the percentage of sentences a method can in fact classify as positive or negative. For instance, suppose that Emoticons' method can classify only 10\% of the sentences in a dataset, corresponding to the actual percentage of sentences with emoticons. It means that the coverage of this method in this specific dataset is 10\%. Note that, the coverage is quite an important metric for a more complete evaluation in the 2-class experiments. Even though Emoticons presents high accuracy for the classified phrases, it was not able to make a prediction for 90\% of the sentences. More formally, coverage is calculated as the number of total sentences minus the number of undefined sentences, all of this divided by the total of sentences, where the number of undefined sentences includes neutral outputs for 3-class methods.

$$Coverage = \frac{\# Sentences - \# Undefined}{\# Sentences}$$

\subsection*{Comparison Metrics}

Considering the 3-class comparison experiments, we used the traditional Precision, Recall, and F1 measures for the automated classification.

\begin{table}[h]\centering
\small
\begin{tabular}{cc|ccc}
& & \multicolumn{3}{c}{\textit{Predicted}}\\
& & Positive & Neutral & Negative\\
\hline
& Positive & a & b & c  \\
\textit{Actual} & Neutral & d & e & f \\
& Negative & g & h & i \\
\end{tabular}
\end{table}

Each letter in the above table  represents the number of instances  which are actually in class X and predicted as
class Y, where X;Y $\in$ {positive; neutral; negative}.
The recall (R) of a class $X$ is the ratio of the number of elements correctly classified as $X$ to the number of known elements in class $X$. Precision (P)
of a class $X$ is the ratio of the number of elements classified correctly as $X$ to the total predicted as the class $X$.
For example, the precision of the negative class is
computed as: $P(neg) = i/(c+f+i)$; its recall, as: $R(neg) = i/(g+h+i)$; and the $F1$ measure is the harmonic mean  between both precision and recall. In this case, $F1(neg)=\frac{2P(neg)\cdot R(neg)}{P(neg)+R(neg)}$.

We also compute the overall accuracy as: $A = \frac{a+e+i}{a+b+c+d+e+f+g+h+i}$. It considers equally important the correct classification of each sentence, independently of the class, and basically measures the capability of the method to predict the correct output. A  variation of F1, namely, Macro-F1, is
normally reported to evaluate classification effectiveness on skewed datasets.
Macro-F1 values are computed by first calculating F1 values for
each class in isolation, as exemplified above for negative, and then averaging over all classes.  Macro-F1 considers equally important the effectiveness in \textit{each class}, independently of the relative size of the class. Thus, accuracy and Macro-F1  provide complementary assessments of the classification effectiveness. Macro-F1 is especially important when the class distribution is very skewed, to verify the capability of the method to perform well in the
smaller classes.

The described metrics can be easily computed for the 2-class experiments by just removing neutral columns and rows as in the table below.

\begin{table}[h]\centering
\small
\begin{tabular}{cc|ccc}
& & \multicolumn{2}{c}{\textit{Predicted}}\\
& & Positive & Negative\\
\hline
& Positive & a & b \\
\textit{Actual} & Negative & c & d \\
 \\
\end{tabular}
\end{table}

In this case, the precision of positive class is computed as: $P(pos) = a/(a+c)$; its recall as: $R(pos) = a/(a+b)$; while its F1 is $F1(pos)=\frac{2P(pos)\cdot R(pos)}{P(pos)+R(pos)}$

As we have a large number of combinations among the base methods, metrics  and datasets, a global analysis of the performance of all these combinations is not an easy task. We propose a simple but informative measure to assess the overall performance ranking. The Mean Ranking is basically the sum of ranks obtained by a method in each dataset divided by the total number of datasets, as below:
\[MR = \frac{\sum\limits_{j=1}^{nd}ri}{nd}\]
where $nd$ is the number of datasets and $ri$ is the rank of the method for dataset $i$. It is important to notice that the rank was calculated based on Macro F1.

The last evaluation metric we exploit is  the Friedman's Test~\cite{friedman2013}. It allows one to verify whether, in a specific experiment, the observed values are globally similar. We used this test to tell if the methods present similar performance across different datasets. More specifically, suppose that $k$ expert raters evaluated $n$ item -- the question that arises is: are rates provided by judges consistent with each other or do they follow completely different patterns? The application in our context is very similar: the datasets are the judges and the Macro-F1 achieved by a method is the rating from the judges.

The Friedman's Test is applied to rankings. Then, to proceed with this statistical test, we sort the methods in decreasing order of Macro-F1 for each dataset.
More formally, the Friedman's rank test in our experiment is defined as:

\[F_R = (\frac{12}{rc(c+1)}\ {\sum\limits_{j=1}^{c}R^2_j}) -3r(c+1) \]

where

$R^2_j$ = square of the sum of rank positions of method j (j = 1,2,..,c)

$r$ = number of datasets

$c$ = number of methods

As the number of datasets increases, the statistical test can be approximated by using the chi-square distribution with $c-1$ degrees of freedom \cite{jain91}. Then, if the $F_R$ computed value is larger than the critical value for the chi-square distribution the null hypothesis is rejected. This null hypothesis states that ranks obtained per dataset are globally similar. Accordingly, rejecting the null hypothesis means that there are significant differences in the ranks across datasets. It is important to note that, in general, the critical value is obtained with significance level $\alpha = 0.05$. Synthesizing, the null hypothesis should be rejected if $F_R > X^2_\alpha$, where $X^2_\alpha$ is the critical value verified in the chi-square distribution table with $c-1$ degrees of freedom and $\alpha$ equals $0.05$.

\subsection*{Comparing Prediction Performance}

We start the analysis of our experiments by comparing the results of all previously discussed metrics for all datasets. Table~\ref{tab:tabelao2} and Table~\ref{tab:tabelao} present accuracy, precision, and Macro-F1 for all methods considering four datasets for the 2-class and 3-class experiments, respectively. For simplicity, we choose to discuss results only for these datasets as they come from different sources and help us to illustrate the main findings from our analysis. Results for all the other datasets are presented in the Additional File 1. There are many interesting observations we can make from these results, summarized next.

\begin{table*}[htpb]
  \centering
  \tiny
  \caption{2-classes experiments results with 4 datasets}
    \begin{tabular}{c|r|c|c|c|c|c|c|c|c|c}
    \hline
    \multirow{2}[2]{*}{\textbf{Dataset}} & \multicolumn{1}{c|}{\multirow{2}[2]{*}{\textbf{Method}}} & \multirow{2}[2]{*}{\textbf{Accur.}} & \multicolumn{3}{c|}{\textbf{Posit. Sentiment}} & \multicolumn{3}{c|}{\textbf{Negat. Sentiment}} & \multirow{2}[2]{*}{\textbf{MacroF1}} & \multirow{2}[2]{*}{\textbf{Coverage}} \\
          & \multicolumn{1}{c|}{} &       & \multicolumn{1}{c}{\textbf{P}} & \multicolumn{1}{c}{\textbf{R}} & \textbf{F1} & \multicolumn{1}{c}{\textbf{P}} & \multicolumn{1}{c}{\textbf{R}} & \textbf{F1} &  & \\

\hline
 \textbf{} & \multicolumn{1}{l|}{AFINN} & 96.37 & 97.66 & 96.94 & 97.30 & 93.75 & 95.19 & 94.47 & 95.88 & 80.77\\
 \textbf{} & \multicolumn{1}{l|}{ANEW\_SUB} & 81.36 & 80.52 & 96.38 & 87.74 & 85.44 & 47.64 & 61.17 & 74.45 & 93.35\\
 \textbf{} & \multicolumn{1}{l|}{Emolex} & 86.06 & 89.82 & 89.11 & 89.47 & 78.77 & 80.00 & 79.38 & 84.42 & 63.58\\
 \textbf{} & \multicolumn{1}{l|}{Emoticons} & 97.75 & 97.90 & 99.42 & 98.65 & 96.97 & 89.72 & 93.20 & 95.93 & 14.82\\
 \textbf{} & \multicolumn{1}{l|}{Emoticons DS} & 71.04 & 70.61 & 99.90 & 82.74 & 95.83 & 5.43 & 10.28 & 46.51 & 99.09\\
 \textbf{} & \multicolumn{1}{l|}{NRC Hashtag} & 67.37 & 83.76 & 65.43 & 73.47 & 48.17 & 71.69 & 57.62 & 65.55 & 91.94\\
 \textbf{} & \multicolumn{1}{l|}{LIWC07} & 66.47 & 74.46 & 78.81 & 76.58 & 44.20 & 38.31 & 41.04 & 58.81 & 73.93\\
 \textbf{} & \multicolumn{1}{l|}{LIWC15} & 96.44 & 97.09 & 98.04 & 97.56 & 94.68 & 92.23 & 93.44 & 95.50 & 77.05\\
 \textbf{} & \multicolumn{1}{l|}{Opinion Finder} & 78.32 & 93.86 & 71.11 & 80.92 & 63.42 & 91.50 & 74.92 & 77.92 & 41.23\\
 \textbf{} & \multicolumn{1}{l|}{Opinion Lexicon} & 93.45 & 97.03 & 93.14 & 95.04 & 86.93 & 94.11 & 90.38 & 92.71 & 70.64\\
 \textbf{Tweets} & \multicolumn{1}{l|}{PANAS-t} & 90.71 & 96.95 & 88.19 & 92.36 & 82.11 & 95.12 & 88.14 & 90.25 & 5.39\\
 \textbf{\_RND\_II} & \multicolumn{1}{l|}{Pattern.en} & 91.76 & 92.94 & 96.19 & 94.54 & 87.86 & 79.06 & 83.23 & 88.88 & 70.85\\
 \textbf{} & \multicolumn{1}{l|}{SASA} & 70.06 & 82.81 & 72.81 & 77.49 & 49.05 & 63.39 & 55.30 & 66.40 & 63.04\\
 \textbf{} & \multicolumn{1}{l|}{Semantria} & 91.61 & 96.94 & 90.55 & 93.64 & 82.25 & 93.88 & 87.68 & 90.66 & 63.61\\
 \textbf{} & \multicolumn{1}{l|}{SenticNet} & 73.64 & 90.74 & 68.45 & 78.03 & 55.41 & 84.88 & 67.05 & 72.54 & 82.82\\
 \textbf{} & \multicolumn{1}{l|}{Sentiment140} & 94.75 & 97.10 & 95.71 & 96.40 & 88.64 & 92.13 & 90.35 & 93.37 & 49.95\\
 \textbf{} & \multicolumn{1}{l|}{Sentiment140\_L} & 78.05 & 88.68 & 78.31 & 83.17 & 61.32 & 77.47 & 68.45 & 75.81 & 93.28\\
 \textbf{} & \multicolumn{1}{l|}{SentiStrength} & 96.97 & 98.92 & 96.43 & 97.66 & 93.54 & 98.01 & 95.72 & 96.69 & 34.65\\
 \textbf{} & \multicolumn{1}{l|}{SentiWordNet} & 78.57 & 87.88 & 80.91 & 84.25 & 61.09 & 72.87 & 66.46 & 75.36 & 61.49\\
 \textbf{} & \multicolumn{1}{l|}{SO-CAL} & 87.76 & 94.25 & 86.99 & 90.47 & 77.34 & 89.32 & 82.90 & 86.68 & 67.18\\
 \textbf{} & \multicolumn{1}{l|}{Stanford DM} & 60.46 & 94.48 & 44.87 & 60.84 & 44.06 & 94.30 & 60.06 & 60.45 & 88.89\\
 \textbf{} & \multicolumn{1}{l|}{Umigon} & 88.63 & 97.73 & 85.92 & 91.45 & 73.64 & 95.17 & 83.03 & 87.24 & 70.83\\
  \textbf{} & \multicolumn{1}{l|}{USent} & 84.46 & 89.28 & 87.67 & 88.47 & 74.77 & 77.63 & 76.17 & 82.32 & 38.94\\
 \textbf{} & \multicolumn{1}{l|}{VADER} & 99.04 & 99.16 & 99.45 & 99.31 & 98.77 & 98.12 & 98.45 & 98.88 & 94.40\\
\hline
 \textbf{} & \multicolumn{1}{l|}{AFINN} & 84.42 & 80.62 & 91.49 & 85.71 & 89.66 & 77.04 & 82.87 & 84.29 & 76.88\\
 \textbf{} & \multicolumn{1}{l|}{ANEW\_SUB} & 68.05 & 63.08 & 93.18 & 75.23 & 84.62 & 40.74 & 55.00 & 65.11 & 94.15\\
 \textbf{} & \multicolumn{1}{l|}{Emolex} & 79.65 & 76.09 & 88.98 & 82.03 & 85.23 & 69.44 & 76.53 & 79.28 & 62.95\\
 \textbf{} & \multicolumn{1}{l|}{Emoticons} & 85.42 & 80.65 & 96.15 & 87.72 & 94.12 & 72.73 & 82.05 & 84.89 & 13.37\\
 \textbf{} & \multicolumn{1}{l|}{Emoticons DS} & 51.96 & 51.41 & 100.00 & 67.91 & 100.00 & 2.27 & 4.44 & 36.18 & 99.72\\
 \textbf{} & \multicolumn{1}{l|}{NRC Hashtag} & 71.30 & 73.05 & 70.93 & 71.98 & 69.51 & 71.70 & 70.59 & 71.28 & 92.20\\
 \textbf{} & \multicolumn{1}{l|}{LIWC07} & 64.29 & 63.75 & 76.12 & 69.39 & 65.22 & 50.85 & 57.14 & 63.27 & 70.39\\
 \textbf{} & \multicolumn{1}{l|}{LIWC15} & 89.22 & 84.18 & 97.08 & 90.17 & 96.40 & 81.06 & 88.07 & 89.12 & 74.93\\
 \textbf{} & \multicolumn{1}{l|}{Opinion Finder} & 80.77 & 81.16 & 76.71 & 78.87 & 80.46 & 84.34 & 82.35 & 80.61 & 43.45\\
 \textbf{} & \multicolumn{1}{l|}{Opinion Lexicon} & 86.10 & 83.67 & 91.11 & 87.23 & 89.29 & 80.65 & 84.75 & 85.99 & 72.14\\
 \textbf{Tweets} & \multicolumn{1}{l|}{PANAS-t} & 94.12 & 88.89 & 100.00 & 94.12 & 100.00 & 88.89 & 94.12 & 94.12 & 4.74\\
 \textbf{\_STF} & \multicolumn{1}{l|}{Pattern.en} & 79.55 & 74.86 & 94.48 & 83.54 & 90.12 & 61.34 & 73.00 & 78.27 & 73.54\\
 \textbf{} & \multicolumn{1}{l|}{SASA} & 68.52 & 65.65 & 78.90 & 71.67 & 72.94 & 57.94 & 64.58 & 68.12 & 60.17\\
 \textbf{} & \multicolumn{1}{l|}{Semantria} & 88.45 & 89.15 & 88.46 & 88.80 & 87.70 & 88.43 & 88.07 & 88.43 & 69.92\\
 \textbf{} & \multicolumn{1}{l|}{SenticNet} & 70.49 & 71.31 & 63.50 & 67.18 & 69.88 & 76.82 & 73.19 & 70.18 & 80.22\\
 \textbf{} & \multicolumn{1}{l|}{Sentiment140} & 93.29 & 91.36 & 94.87 & 93.08 & 95.18 & 91.86 & 93.49 & 93.29 & 45.68\\
 \textbf{} & \multicolumn{1}{l|}{Sentiment140\_L} & 79.12 & 81.48 & 76.30 & 78.81 & 76.97 & 82.04 & 79.42 & 79.11 & 94.71\\
 \textbf{} & \multicolumn{1}{l|}{SentiStrength} & 95.33 & 95.18 & 96.34 & 95.76 & 95.52 & 94.12 & 94.81 & 95.29 & 41.78\\
 \textbf{} & \multicolumn{1}{l|}{SentiWordNet} & 72.99 & 73.17 & 78.95 & 75.95 & 72.73 & 65.98 & 69.19 & 72.57 & 58.77\\
 \textbf{} & \multicolumn{1}{l|}{SO-CAL} & 87.36 & 82.89 & 93.33 & 87.80 & 92.80 & 81.69 & 86.89 & 87.35 & 77.16\\
 \textbf{} & \multicolumn{1}{l|}{Stanford DM} & 66.56 & 87.69 & 36.31 & 51.35 & 61.24 & 95.18 & 74.53 & 62.94 & 89.97\\
 \textbf{} & \multicolumn{1}{l|}{Umigon} & 86.99 & 91.73 & 81.88 & 86.52 & 83.02 & 92.31 & 87.42 & 86.97 & 81.34\\
 \textbf{} & \multicolumn{1}{l|}{USent} & 73.21 & 69.35 & 82.69 & 75.44 & 78.82 & 63.81 & 70.53 & 72.98 & 58.22\\
 \textbf{} & \multicolumn{1}{l|}{VADER} & 84.44 & 80.23 & 92.21 & 85.80 & 90.40 & 76.35 & 82.78 & 84.29 & 84.12\\
\hline
 \textbf{} & \multicolumn{1}{l|}{AFINN} & 70.94 & 47.01 & 81.82 & 59.72 & 91.17 & 67.05 & 77.27 & 68.49 & 74.81\\
 \textbf{} & \multicolumn{1}{l|}{ANEW\_SUB} & 43.25 & 30.98 & 92.31 & 46.39 & 90.13 & 25.46 & 39.71 & 43.05 & 93.73\\
 \textbf{} & \multicolumn{1}{l|}{Emolex} & 61.71 & 34.60 & 75.83 & 47.52 & 88.93 & 57.53 & 69.87 & 58.69 & 67.14\\
 \textbf{} & \multicolumn{1}{l|}{Emoticons} & 73.08 & 72.22 & 86.67 & 78.79 & 75.00 & 54.55 & 63.16 & 70.97 & 3.32\\
 \textbf{} & \multicolumn{1}{l|}{Emoticons DS} & 28.24 & 27.30 & 100.00 & 42.89 & 100.00 & 1.77 & 3.48 & 23.19 & 98.72\\
 \textbf{} & \multicolumn{1}{l|}{NRC Hashtag} & 74.69 & 51.01 & 40.64 & 45.24 & 80.80 & 86.48 & 83.54 & 64.39 & 92.97\\
 \textbf{} & \multicolumn{1}{l|}{LIWC07} & 46.15 & 27.44 & 58.40 & 37.34 & 72.49 & 41.52 & 52.79 & 45.07 & 58.18\\
 \textbf{} & \multicolumn{1}{l|}{LIWC15} & 70.67 & 49.81 & 90.91 & 64.36 & 94.35 & 62.36 & 75.09 & 69.72 & 62.79\\
 \textbf{} & \multicolumn{1}{l|}{Opinion Finder} & 71.14 & 43.04 & 64.76 & 51.71 & 86.88 & 73.13 & 79.42 & 65.56 & 56.27\\
 \textbf{} & \multicolumn{1}{l|}{Opinion Lexicon} & 71.82 & 47.45 & 86.43 & 61.27 & 93.40 & 66.75 & 77.86 & 69.56 & 69.44\\
 \textbf{Comments} & \multicolumn{1}{l|}{PANAS-t} & 68.00 & 12.50 & 50.00 & 20.00 & 94.12 & 69.57 & 80.00 & 50.00 & 3.20\\
 \textbf{\_Digg} & \multicolumn{1}{l|}{Pattern.en} & 60.05 & 43.73 & 92.14 & 59.31 & 92.57 & 45.21 & 60.75 & 60.03 & 56.65\\
 \textbf{} & \multicolumn{1}{l|}{SASA} & 65.54 & 40.26 & 66.91 & 50.27 & 84.82 & 65.06 & 73.64 & 61.95 & 68.29\\
 \textbf{} & \multicolumn{1}{l|}{Semantria} & 82.46 & 62.72 & 88.33 & 73.36 & 94.81 & 80.25 & 86.93 & 80.14 & 56.14\\
 \textbf{} & \multicolumn{1}{l|}{SenticNet} & 69.40 & 46.30 & 72.46 & 56.50 & 86.77 & 68.25 & 76.40 & 66.45 & 96.55\\
 \textbf{} & \multicolumn{1}{l|}{Sentiment140} & 85.06 & 62.50 & 78.95 & 69.77 & 93.65 & 86.76 & 90.08 & 79.92 & 33.38\\
 \textbf{} & \multicolumn{1}{l|}{Sentiment140\_L} & 67.76 & 42.07 & 73.45 & 53.50 & 88.01 & 65.84 & 75.33 & 64.41 & 89.64\\
 \textbf{} & \multicolumn{1}{l|}{SentiStrength} & 92.09 & 78.69 & 92.31 & 84.96 & 97.40 & 92.02 & 94.64 & 89.80 & 27.49\\
 \textbf{} & \multicolumn{1}{l|}{SentiWordNet} & 62.17 & 36.86 & 77.68 & 50.00 & 88.84 & 57.18 & 69.58 & 59.79 & 58.82\\
 \textbf{} & \multicolumn{1}{l|}{SO-CAL} & 76.55 & 52.86 & 77.08 & 62.71 & 90.65 & 76.37 & 82.90 & 72.81 & 71.99\\
 \textbf{} & \multicolumn{1}{l|}{Stanford DM} & 69.16 & 35.29 & 20.27 & 25.75 & 75.21 & 86.68 & 80.54 & 53.15 & 78.90\\
 \textbf{} & \multicolumn{1}{l|}{Umigon} & 83.37 & 66.22 & 75.38 & 70.50 & 90.72 & 86.23 & 88.42 & 79.46 & 63.04\\
 \textbf{} & \multicolumn{1}{l|}{USent} & 55.98 & 36.06 & 80.65 & 49.83 & 86.67 & 46.80 & 60.78 & 55.31 & 43.86\\ 
 \textbf{} & \multicolumn{1}{l|}{VADER} & 69.05 & 45.48 & 85.88 & 59.47 & 92.55 & 63.00 & 74.97 & 67.22 & 82.23\\
\hline
 \textbf{} & \multicolumn{1}{l|}{AFINN} & 66.56 & 23.08 & 81.08 & 35.93 & 96.32 & 64.66 & 77.38 & 56.65 & 85.11\\
 \textbf{} & \multicolumn{1}{l|}{ANEW\_SUB} & 31.37 & 15.48 & 95.79 & 26.65 & 97.18 & 21.73 & 35.52 & 31.08 & 97.07\\
 \textbf{} & \multicolumn{1}{l|}{Emolex} & 59.64 & 21.52 & 89.04 & 34.67 & 97.38 & 55.62 & 70.80 & 52.73 & 80.72\\
 \textbf{} & \multicolumn{1}{l|}{Emoticons} & 33.33 & 0.00 & 0.00 & 0.00 & 100.00 & 33.33 & 50.00 & 25.00 & 0.40\\
 \textbf{} & \multicolumn{1}{l|}{Emoticons DS} & 13.33 & 13.10 & 100.00 & 23.17 & 100.00 & 0.31 & 0.61 & 11.89 & 99.73\\
 \textbf{} & \multicolumn{1}{l|}{NRC Hashtag} & 84.45 & 33.33 & 25.27 & 28.75 & 89.76 & 92.83 & 91.27 & 60.01 & 97.47\\
 \textbf{} & \multicolumn{1}{l|}{LIWC07} & 50.10 & 15.38 & 58.33 & 24.35 & 88.00 & 48.78 & 62.77 & 43.56 & 69.55\\
 \textbf{} & \multicolumn{1}{l|}{LIWC15} & 63.21 & 25.86 & 90.67 & 40.24 & 97.55 & 58.86 & 73.42 & 56.83 & 73.01\\
 \textbf{} & \multicolumn{1}{l|}{Opinion Finder} & 74.43 & 21.74 & 62.50 & 32.26 & 94.93 & 75.72 & 84.24 & 58.25 & 76.46\\
 \textbf{} & \multicolumn{1}{l|}{Opinion Lexicon} & 74.14 & 29.81 & 84.93 & 44.13 & 97.24 & 72.66 & 83.17 & 63.65 & 80.72\\
 \textbf{Comments} & \multicolumn{1}{l|}{PANAS-t} & 58.73 & 20.00 & 75.00 & 31.58 & 93.94 & 56.36 & 70.45 & 51.02 & 8.38\\
 \textbf{\_BBC} & \multicolumn{1}{l|}{Pattern.en} & 41.75 & 19.73 & 93.55 & 32.58 & 96.61 & 32.57 & 48.72 & 40.65 & 54.79\\
 \textbf{} & \multicolumn{1}{l|}{SASA} & 61.61 & 23.50 & 66.20 & 34.69 & 90.80 & 60.77 & 72.81 & 53.75 & 61.30\\
 \textbf{} & \multicolumn{1}{l|}{Semantria} & 83.43 & 40.00 & 84.75 & 54.35 & 97.64 & 83.26 & 89.88 & 72.11 & 67.42\\
 \textbf{} & \multicolumn{1}{l|}{SenticNet} & 66.07 & 24.44 & 74.16 & 36.77 & 94.24 & 64.83 & 76.81 & 56.79 & 88.96\\
 \textbf{} & \multicolumn{1}{l|}{Sentiment140} & 68.51 & 24.00 & 69.77 & 35.71 & 94.04 & 68.33 & 79.15 & 57.43 & 45.61\\
 \textbf{} & \multicolumn{1}{l|}{Sentiment140\_L} & 56.85 & 18.52 & 69.15 & 29.21 & 92.35 & 55.03 & 68.97 & 49.09 & 97.07\\
 \textbf{} & \multicolumn{1}{l|}{SentiStrength} & 93.93 & 64.29 & 78.26 & 70.59 & 97.72 & 95.54 & 96.61 & 83.60 & 32.85\\
 \textbf{} & \multicolumn{1}{l|}{SentiWordNet} & 57.49 & 20.00 & 88.06 & 32.60 & 97.13 & 53.45 & 68.96 & 50.78 & 76.33\\
 \textbf{} & \multicolumn{1}{l|}{SO-CAL} & 75.28 & 28.93 & 80.28 & 42.54 & 96.71 & 74.64 & 84.25 & 63.40 & 82.85\\
 \textbf{} & \multicolumn{1}{l|}{Stanford DM} & 89.45 & 63.16 & 40.91 & 49.66 & 91.81 & 96.52 & 94.11 & 71.88 & 92.02\\
 \textbf{} & \multicolumn{1}{l|}{Umigon} & 79.37 & 39.13 & 61.02 & 47.68 & 92.10 & 82.72 & 87.15 & 67.42 & 50.93\\
 \textbf{} & \multicolumn{1}{l|}{USent} & 52.60 & 18.33 & 80.49 & 29.86 & 94.56 & 48.60 & 64.20 & 47.03 & 43.48\\ 
 \textbf{} & \multicolumn{1}{l|}{VADER} & 62.76 & 22.68 & 85.54 & 35.86 & 96.75 & 59.60 & 73.76 & 54.81 & 90.69\\
\hline   
    \end{tabular}%
  \label{tab:tabelao2}%
\end{table*}%

\begin{table*}[htpb]
  \centering
  \tiny
  \caption{3-classes experiments results with 4 datasets}
    \begin{tabular}{c|r|c|c|c|c|c|c|c|c|c|c|c}
    \hline
    \multirow{2}[2]{*}{\textbf{Dataset}} & \multicolumn{1}{c|}{\multirow{2}[2]{*}{\textbf{Method}}} & \multirow{2}[2]{*}{\textbf{Accur.}} & \multicolumn{3}{c|}{\textbf{Posit. Sentiment }} & \multicolumn{3}{c|}{\textbf{Negat. Sentiment}} & \multicolumn{3}{c|}{\textbf{Neut. Sentiment}} & \multirow{2}[2]{*}{\textbf{MacroF1}} \\
          & \multicolumn{1}{c|}{} &       & \multicolumn{1}{c}{\textbf{P}} & \multicolumn{1}{c}{\textbf{R}} & \textbf{F1} & \multicolumn{1}{c}{\textbf{P}} & \multicolumn{1}{c}{\textbf{R}} & \textbf{F1} & \multicolumn{1}{c}{\textbf{P}} & \multicolumn{1}{c}{\textbf{R}} & \textbf{F1} &  \\

 \hline
\textbf{} & \multicolumn{1}{l|}{AFINN} & 62.36 & 61.10 & 70.09 & 65.28 & 44.08 & 55.56 & 49.15 & 71.43 & 58.57 & 64.37 & 59.60\\
\textbf{} & \multicolumn{1}{l|}{ANEW\_SUB} & 39.51 & 38.79 & 96.31 & 55.31 & 43.50 & 23.18 & 30.24 & 57.38 & 2.31 & 4.45 & 30.00\\
\textbf{} & \multicolumn{1}{l|}{Emolex} & 48.74 & 48.15 & 62.71 & 54.47 & 31.27 & 38.59 & 34.55 & 57.90 & 41.30 & 48.21 & 45.74\\
\textbf{} & \multicolumn{1}{l|}{Emoticons} & 52.88 & 72.83 & 11.34 & 19.62 & 55.56 & 5.38 & 9.80 & 34.05 & 96.53 & 50.34 & 26.59\\
\textbf{} & \multicolumn{1}{l|}{Emoticons DS} & 36.59 & 36.55 & 100.00 & 53.53 & 75.00 & 0.36 & 0.71 & 100.00 & 0.03 & 0.07 & 18.10\\
\textbf{} & \multicolumn{1}{l|}{NRC Hashtag} & 36.95 & 42.04 & 75.03 & 53.88 & 24.57 & 56.03 & 34.16 & 53.33 & 3.70 & 6.92 & 31.65\\
\textbf{} & \multicolumn{1}{l|}{LIWC07} & 39.54 & 36.52 & 42.33 & 39.21 & 15.14 & 13.02 & 14.00 & 48.64 & 44.83 & 46.66 & 33.29\\
\textbf{} & \multicolumn{1}{l|}{LIWC15} & 62.56 & 59.77 & 71.03 & 64.91 & 49.04 & 42.65 & 45.62 & 68.90 & 61.84 & 65.18 & 58.57\\
\textbf{} & \multicolumn{1}{l|}{Opinion Finder} & 57.63 & 67.57 & 27.94 & 39.53 & 40.75 & 33.69 & 36.89 & 58.20 & 86.06 & 69.44 & 48.62\\
\textbf{} & \multicolumn{1}{l|}{Opinion Lexicon} & 60.37 & 62.09 & 62.71 & 62.40 & 41.19 & 52.81 & 46.28 & 66.41 & 60.75 & 63.46 & 57.38\\
\textbf{Tweets} & \multicolumn{1}{l|}{PANAS-t} & 53.08 & 90.95 & 9.04 & 16.45 & 51.56 & 3.94 & 7.33 & 51.65 & 99.01 & 67.89 & 30.55\\
\textbf{\_Semeval} & \multicolumn{1}{l|}{Pattern.en} & 57.99 & 57.97 & 68.74 & 62.89 & 34.83 & 35.24 & 35.04 & 65.55 & 56.39 & 60.63 & 52.85\\
\textbf{} & \multicolumn{1}{l|}{SASA} & 50.63 & 46.34 & 47.77 & 47.04 & 33.07 & 20.31 & 25.17 & 56.39 & 61.12 & 58.66 & 43.62\\
\textbf{} & \multicolumn{1}{l|}{Semantria} & 61.54 & 67.28 & 57.35 & 61.92 & 39.57 & 52.81 & 45.24 & 65.98 & 67.03 & 66.50 & 57.89\\
\textbf{} & \multicolumn{1}{l|}{SenticNet} & 49.68 & 51.85 & 1.26 & 2.46 & 29.79 & 1.67 & 3.17 & 49.82 & 98.51 & 66.17 & 23.93\\
\textbf{} & \multicolumn{1}{l|}{Sentiment140} & 60.42 & 63.87 & 51.37 & 56.94 & 50.96 & 37.87 & 43.45 & 60.35 & 73.31 & 66.20 & 55.53\\
\textbf{} & \multicolumn{1}{l|}{Sentiment140\_L} & 39.44 & 43.52 & 74.72 & 55.00 & 27.67 & 65.35 & 38.88 & 65.87 & 6.38 & 11.63 & 35.17\\
\textbf{} & \multicolumn{1}{l|}{SentiStrength} & 57.83 & 78.01 & 27.13 & 40.25 & 47.80 & 23.42 & 31.44 & 55.49 & 89.89 & 68.62 & 46.77\\
\textbf{} & \multicolumn{1}{l|}{SentiWordNet} & 48.33 & 55.54 & 53.44 & 54.47 & 19.67 & 37.51 & 25.81 & 61.22 & 47.57 & 53.54 & 44.61\\
\textbf{} & \multicolumn{1}{l|}{SO-CAL} & 58.83 & 58.89 & 59.02 & 58.95 & 40.39 & 54.24 & 46.30 & 39.89 & 59.96 & 47.91 & 51.05\\
\textbf{} & \multicolumn{1}{l|}{Stanford DM} & 22.54 & 72.14 & 18.17 & 29.03 & 14.92 & 90.56 & 25.61 & 47.19 & 6.94 & 12.10 & 22.25\\
\textbf{} & \multicolumn{1}{l|}{Umigon} & 65.88 & 75.18 & 56.14 & 64.28 & 39.66 & 55.91 & 46.41 & 70.65 & 75.78 & 73.13 & 61.27\\
\textbf{} & \multicolumn{1}{l|}{USent} & 52.13 & 49.86 & 32.88 & 39.63 & 39.96 & 22.82 & 29.05 & 54.33 & 74.36 & 62.79 & 43.82\\s
\textbf{} & \multicolumn{1}{l|}{VADER} & 60.21 & 56.46 & 79.04 & 65.87 & 44.30 & 59.02 & 50.61 & 76.02 & 46.71 & 57.87 & 58.12\\
\hline
\textbf{} & \multicolumn{1}{l|}{AFINN} & 64.41 & 40.81 & 72.12 & 52.13 & 49.67 & 62.50 & 55.35 & 85.95 & 62.54 & 72.40 & 59.96\\
\textbf{} & \multicolumn{1}{l|}{ANEW\_SUB} & 28.03 & 21.89 & 92.29 & 35.38 & 44.30 & 34.22 & 38.61 & 74.82 & 8.18 & 14.74 & 29.58\\
\textbf{} & \multicolumn{1}{l|}{Emolex} & 54.76 & 31.67 & 59.95 & 41.44 & 40.14 & 47.54 & 43.53 & 77.48 & 54.64 & 64.08 & 49.68\\
\textbf{} & \multicolumn{1}{l|}{Emoticons} & 70.22 & 70.06 & 16.78 & 27.07 & 65.62 & 8.61 & 15.22 & 41.29 & 97.56 & 58.02 & 33.44\\
\textbf{} & \multicolumn{1}{l|}{Emoticons DS} & 20.34 & 19.78 & 99.46 & 33.00 & 62.07 & 3.69 & 6.96 & 53.85 & 0.55 & 1.09 & 13.68\\
\textbf{} & \multicolumn{1}{l|}{NRC Hashtag} & 30.47 & 28.25 & 77.40 & 41.39 & 24.18 & 72.54 & 36.27 & 79.08 & 8.77 & 15.78 & 31.15\\
\textbf{} & \multicolumn{1}{l|}{LIWC} & 46.88 & 21.85 & 38.43 & 27.86 & 19.18 & 18.24 & 18.70 & 69.51 & 54.83 & 61.31 & 35.95\\
\textbf{} & \multicolumn{1}{l|}{LIWC15} & 67.75 & 44.78 & 78.35 & 56.99 & 57.49 & 57.38 & 57.44 & 85.18 & 66.67 & 74.80 & 63.07\\
\textbf{} & \multicolumn{1}{l|}{Opinion Finder} & 71.55 & 57.48 & 32.75 & 41.72 & 49.85 & 34.63 & 40.87 & 75.95 & 89.90 & 82.34 & 54.98\\
\textbf{} & \multicolumn{1}{l|}{Opinion Lexicon} & 63.86 & 40.65 & 66.17 & 50.36 & 48.84 & 56.15 & 52.24 & 81.96 & 64.66 & 72.29 & 58.30\\
\textbf{Tweets} & \multicolumn{1}{l|}{PANAS-t} & 68.79 & 79.49 & 8.39 & 15.18 & 48.57 & 3.48 & 6.50 & 68.75 & 98.86 & 81.10 & 34.26\\
\textbf{\_RND\_III} & \multicolumn{1}{l|}{Pattern.en} & 59.56 & 36.20 & 77.00 & 49.24 & 52.87 & 45.29 & 48.79 & 81.75 & 57.23 & 67.33 & 55.12\\
\textbf{} & \multicolumn{1}{l|}{SASA} & 55.37 & 29.42 & 54.53 & 38.22 & 42.46 & 47.34 & 44.77 & 78.30 & 57.15 & 66.08 & 49.69\\
\textbf{} & \multicolumn{1}{l|}{Semantria} & 68.89 & 48.86 & 63.73 & 55.31 & 49.82 & 55.53 & 52.52 & 82.02 & 72.96 & 77.22 & 61.68\\
\textbf{} & \multicolumn{1}{l|}{SenticNet} & 29.97 & 31.08 & 74.83 & 43.92 & 20.98 & 73.98 & 32.68 & 79.70 & 8.49 & 15.35 & 30.65\\
\textbf{} & \multicolumn{1}{l|}{Sentiment140} & 76.40 & 64.42 & 51.69 & 57.36 & 74.75 & 45.49 & 56.56 & 79.04 & 89.50 & 83.94 & 65.95\\
\textbf{} & \multicolumn{1}{l|}{Sentiment140\_L} & 31.32 & 25.83 & 77.13 & 38.70 & 30.05 & 78.69 & 43.49 & 79.37 & 8.92 & 16.04 & 32.74\\
\textbf{} & \multicolumn{1}{l|}{SentiStrength} & 73.80 & 70.94 & 41.95 & 52.72 & 57.53 & 25.82 & 35.64 & 75.35 & 92.26 & 82.95 & 57.10\\
\textbf{} & \multicolumn{1}{l|}{SentiWordNet} & 55.85 & 37.42 & 58.19 & 45.55 & 24.04 & 35.86 & 28.78 & 79.25 & 59.00 & 67.64 & 47.33\\
\textbf{} & \multicolumn{1}{l|}{SO-CAL} & 66.51 & 43.06 & 68.88 & 52.99 & 51.84 & 60.66 & 55.90 & 45.77 & 66.94 & 54.37 & 54.42\\
\textbf{} & \multicolumn{1}{l|}{Stanford DM} & 31.90 & 64.48 & 38.57 & 48.26 & 15.58 & 85.04 & 26.33 & 75.64 & 19.77 & 31.35 & 35.32\\
\textbf{} & \multicolumn{1}{l|}{Umigon} & 74.12 & 57.67 & 70.23 & 63.33 & 48.83 & 68.44 & 57.00 & 88.80 & 76.34 & 82.10 & 67.47\\
\textbf{} & \multicolumn{1}{l|}{USent} & 66.06 & 40.60 & 36.81 & 38.61 & 44.87 & 28.69 & 35.00 & 74.54 & 81.72 & 77.97 & 50.53\\
\textbf{} & \multicolumn{1}{l|}{VADER} & 60.14 & 37.69 & 81.60 & 51.56 & 48.56 & 65.57 & 55.80 & 88.96 & 52.87 & 66.32 & 57.89\\
\hline
\textbf{} & \multicolumn{1}{l|}{AFINN} & 50.10 & 16.22 & 60.61 & 25.59 & 82.62 & 56.05 & 66.79 & 40.11 & 30.24 & 34.48 & 42.29\\
\textbf{} & \multicolumn{1}{l|}{ANEW\_SUB} & 24.30 & 11.38 & 91.92 & 20.24 & 84.15 & 21.13 & 33.78 & 38.89 & 5.65 & 9.86 & 21.30\\
\textbf{} & \multicolumn{1}{l|}{Emolex} & 44.10 & 15.51 & 65.66 & 25.10 & 83.19 & 45.48 & 58.81 & 35.27 & 31.85 & 33.47 & 39.13\\
\textbf{} & \multicolumn{1}{l|}{Emoticons} & 24.60 & 0.00 & 0.00 & 0.00 & 33.33 & 0.15 & 0.30 & 19.77 & 98.79 & 32.95 & 11.09\\
\textbf{} & \multicolumn{1}{l|}{Emoticons DS} & 10.00 & 9.85 & 98.99 & 17.92 & 66.67 & 0.31 & 0.61 & 0.00 & 0.00 & 0.00 & 6.18\\
\textbf{} & \multicolumn{1}{l|}{NRC Hashtag} & 64.00 & 20.72 & 23.23 & 21.90 & 70.20 & 91.27 & 79.36 & 52.50 & 8.47 & 14.58 & 38.62\\
\textbf{} & \multicolumn{1}{l|}{LIWC07} & 33.00 & 11.11 & 42.42 & 17.61 & 67.69 & 33.69 & 44.99 & 22.90 & 27.42 & 24.95 & 29.18\\
\textbf{} & \multicolumn{1}{l|}{LIWC15} & 43.70 & 17.94 & 68.69 & 28.45 & 85.06 & 42.73 & 56.88 & 30.72 & 36.29 & 33.27 & 39.53\\
\textbf{} & \multicolumn{1}{l|}{Opinion Finder} & 51.80 & 14.96 & 35.35 & 21.02 & 78.76 & 60.18 & 68.23 & 33.71 & 36.29 & 34.95 & 41.40\\
\textbf{} & \multicolumn{1}{l|}{Opinion Lexicon} & 55.00 & 20.67 & 62.63 & 31.08 & 85.27 & 59.42 & 70.04 & 40.82 & 40.32 & 40.57 & 47.23\\
\textbf{Comments} & \multicolumn{1}{l|}{PANAS-t} & 27.10 & 16.67 & 6.06 & 8.89 & 75.61 & 4.75 & 8.93 & 25.35 & 94.35 & 39.97 & 19.26\\
\textbf{\_BBC} & \multicolumn{1}{l|}{Pattern.en} & 28.70 & 14.25 & 58.59 & 22.92 & 82.61 & 17.46 & 28.82 & 25.27 & 46.37 & 32.72 & 28.16\\
\textbf{} & \multicolumn{1}{l|}{SASA} & 38.20 & 17.03 & 47.47 & 25.07 & 70.75 & 36.29 & 47.98 & 25.19 & 39.52 & 30.77 & 34.60\\
\textbf{} & \multicolumn{1}{l|}{Semantria} & 56.00 & 28.90 & 50.51 & 36.76 & 83.82 & 57.12 & 67.94 & 35.86 & 55.24 & 43.49 & 49.40\\
\textbf{} & \multicolumn{1}{l|}{SenticNet} & 47.10 & 17.74 & 66.67 & 28.03 & 72.87 & 57.58 & 64.33 & 25.89 & 11.69 & 16.11 & 36.16\\
\textbf{} & \multicolumn{1}{l|}{Sentiment140} & 40.00 & 17.75 & 30.30 & 22.39 & 79.77 & 31.39 & 45.05 & 28.75 & 66.53 & 40.15 & 35.86\\
\textbf{} & \multicolumn{1}{l|}{Sentiment140\_L} & 43.10 & 13.32 & 65.66 & 22.15 & 73.84 & 53.60 & 62.11 & 42.11 & 6.45 & 11.19 & 31.82\\
\textbf{} & \multicolumn{1}{l|}{SentiStrength} & 44.20 & 47.37 & 18.18 & 26.28 & 86.64 & 32.77 & 47.56 & 29.37 & 84.68 & 43.61 & 39.15\\
\textbf{} & \multicolumn{1}{l|}{SentiWordNet} & 42.40 & 14.90 & 59.60 & 23.84 & 81.63 & 41.50 & 55.03 & 34.56 & 37.90 & 36.15 & 38.34\\
\textbf{} & \multicolumn{1}{l|}{SO-CAL} & 55.50 & 20.88 & 57.58 & 30.65 & 80.47 & 63.09 & 70.73 & 28.57 & 34.68 & 31.33 & 44.23\\
\textbf{} & \multicolumn{1}{l|}{Stanford DM} & 65.50 & 43.37 & 36.36 & 39.56 & 71.01 & 89.28 & 79.10 & 37.50 & 14.52 & 20.93 & 46.53\\
\textbf{} & \multicolumn{1}{l|}{Umigon} & 45.70 & 28.35 & 36.36 & 31.86 & 76.35 & 41.04 & 53.39 & 29.31 & 61.69 & 39.74 & 41.66\\
\textbf{} & \multicolumn{1}{l|}{USent} & 33.80 & 13.75 & 33.33 & 19.47 & 82.25 & 21.29 & 33.82 & 28.09 & 66.94 & 39.57 & 30.95\\
\textbf{} & \multicolumn{1}{l|}{VADER} & 49.40 & 16.36 & 71.72 & 26.64 & 83.02 & 54.67 & 65.93 & 48.53 & 26.61 & 34.38 & 42.31\\
\hline
\textbf{} & \multicolumn{1}{l|}{AFINN} & 42.45 & 64.81 & 41.79 & 50.81 & 80.29 & 39.82 & 53.24 & 7.89 & 77.87 & 14.32 & 39.46\\
\textbf{} & \multicolumn{1}{l|}{ANEW\_SUB} & 51.12 & 48.35 & 88.57 & 62.55 & 79.65 & 24.69 & 37.69 & 7.92 & 9.84 & 8.78 & 36.34\\
\textbf{} & \multicolumn{1}{l|}{Emolex} & 42.97 & 55.12 & 53.72 & 54.41 & 75.35 & 33.33 & 46.22 & 7.22 & 54.10 & 12.74 & 37.79\\
\textbf{} & \multicolumn{1}{l|}{Emoticons} & 4.68 & 0.00 & 0.00 & 0.00 & 0.00 & 0.00 & 0.00 & 4.47 & 99.59 & 8.56 & 2.85\\
\textbf{} & \multicolumn{1}{l|}{Emoticons DS} & 42.58 & 42.55 & 99.77 & 59.66 & 78.57 & 0.40 & 0.80 & 0.00 & 0.00 & 0.00 & 20.15\\
\textbf{} & \multicolumn{1}{l|}{NRC Hashtag} & 54.84 & 55.38 & 45.74 & 50.10 & 61.55 & 65.68 & 63.55 & 8.33 & 15.16 & 10.76 & 41.47\\
\textbf{} & \multicolumn{1}{l|}{LIWC07} & 24.35 & 42.88 & 27.72 & 33.67 & 53.42 & 19.07 & 28.11 & 4.67 & 53.28 & 8.58 & 23.45\\
\textbf{} & \multicolumn{1}{l|}{LIWC15} & 36.49 & 65.29 & 40.29 & 49.83 & 81.50 & 29.25 & 43.05 & 7.17 & 83.61 & 13.20 & 35.36\\
\textbf{} & \multicolumn{1}{l|}{Opinion Finder} & 29.38 & 68.77 & 18.78 & 29.51 & 76.52 & 32.68 & 45.80 & 6.29 & 88.11 & 11.75 & 29.02\\
\textbf{} & \multicolumn{1}{l|}{Opinion Lexicon} & 44.57 & 65.95 & 43.15 & 52.17 & 79.81 & 43.11 & 55.98 & 7.94 & 73.77 & 14.34 & 40.83\\
\textbf{Comments} & \multicolumn{1}{l|}{PANAS-t} & 5.88 & 69.23 & 1.23 & 2.41 & 62.07 & 1.31 & 2.57 & 4.75 & 99.18 & 9.07 & 4.68\\
\textbf{\_NYT} & \multicolumn{1}{l|}{Pattern.en} & 31.60 & 55.23 & 45.05 & 49.63 & 72.80 & 17.76 & 28.55 & 5.88 & 65.57 & 10.79 & 29.66\\
\textbf{} & \multicolumn{1}{l|}{SASA} & 30.04 & 49.92 & 30.13 & 37.58 & 59.11 & 27.21 & 37.26 & 5.74 & 61.07 & 10.49 & 28.44\\
\textbf{} & \multicolumn{1}{l|}{Semantria} & 44.59 & 70.60 & 41.83 & 52.54 & 80.54 & 44.24 & 57.11 & 7.53 & 73.36 & 13.65 & 41.10\\
\textbf{} & \multicolumn{1}{l|}{SenticNet} & 61.85 & 58.19 & 59.48 & 58.83 & 65.01 & 69.26 & 67.07 & 0.00 & 0.00 & 0.00 & 41.97\\
\textbf{} & \multicolumn{1}{l|}{Sentiment140} & 13.58 & 77.32 & 6.81 & 12.51 & 75.40 & 11.96 & 20.65 & 4.98 & 93.03 & 9.45 & 14.20\\
\textbf{} & \multicolumn{1}{l|}{Sentiment140\_L} & 54.61 & 54.72 & 59.12 & 56.84 & 67.00 & 54.41 & 60.05 & 6.70 & 15.98 & 9.44 & 42.11\\
\textbf{} & \multicolumn{1}{l|}{SentiStrength} & 18.17 & 78.51 & 8.62 & 15.54 & 81.12 & 18.96 & 30.74 & 5.41 & 95.49 & 10.24 & 18.84\\
\textbf{} & \multicolumn{1}{l|}{SentiWordNet} & 32.20 & 57.35 & 34.53 & 43.10 & 70.31 & 26.95 & 38.97 & 6.08 & 70.08 & 11.19 & 31.09\\
\textbf{} & \multicolumn{1}{l|}{SO-CAL} & 50.79 & 64.36 & 51.13 & 56.99 & 77.25 & 49.16 & 60.08 & 8.68 & 65.98 & 15.34 & 44.14\\
\textbf{} & \multicolumn{1}{l|}{Stanford DM} & 51.93 & 73.39 & 21.14 & 32.83 & 59.48 & 77.90 & 67.46 & 9.65 & 38.11 & 15.40 & 38.56\\
\textbf{} & \multicolumn{1}{l|}{Umigon} & 24.08 & 68.76 & 16.38 & 26.46 & 68.78 & 24.51 & 36.14 & 5.88 & 88.93 & 11.04 & 24.54\\
\textbf{} & \multicolumn{1}{l|}{USent} & 27.44 & 56.61 & 28.95 & 38.31 & 77.69 & 21.59 & 33.79 & 5.88 & 79.51 & 10.94 & 27.68\\
\textbf{} & \multicolumn{1}{l|}{VADER} & 48.03 & 62.67 & 51.63 & 56.62 & 79.91 & 43.07 & 55.97 & 9.18 & 71.31 & 16.26 & 42.95\\
    \hline
    \end{tabular}%
  \label{tab:tabelao}%
\end{table*}%

\if 0
 \green{Note that the methods identified by LIWC and LIWC\_V represent the results from original LIWC execution and LIWC Lexicon executed with VADER Heuristics, respectively. This is the reason for twenty-four entries in the Table instead of only twenty-three.} 
\fi

%In some cases methods obtain results worse than a random baseline (i.e. a method that would randomly choose among positive, neutral, or negative as output). This usually happens when a method is biased towards one or more classes. As an example, a naive approach like Emoticons showed to be a good method for detecting positive and negative messages when the input data has emoticons, which does not happen in many cases in our datasets. It considers most of the instances as neutral, as the majority of the messages do not have emoticons, leading to an overall bad performance for most of the datasets.

%However, we note that this bias can be used to construct ensemble approaches. For example, when emoticons position itself towards a positive or negative classification, it should be highly considered as it is usually correct. This can clearly be extended to other methods which showed a similar kind of bias.

\noindent
\textbf{Methods prediction performance varies considerably from one dataset to another}:
 %\if 0 By analyzing Table~\ref{tab:tabelao2} we can note that VADER works well for Tweets\_RND\_I and Tweets\_STF, appearing among the top 3 methods, but it presents poor performance in Tweets\_SAN and Comments\_BBC, achieving the ninth and eleventh place, respectively \fi
First, we note the same social media text can be interpreted very differently depending on the choice of a sentiment method. Overall, we note that all the methods yielded with large variations  across the different datasets. By analyzing Table~\ref{tab:tabelao2} we can note that VADER works well for Tweets\_RND\_II, appearing in the first place, but it presents poor performance in Tweets\_STF, Comments\_BBC,  and Comments\_DIGG, achieving the eleventh, thirteenth and tenth place respectively.  Although the first two datasets contain tweets, they belong to different contexts, which  affects the performance of some methods like VADER. Another important aspect to be analyzed in this Table is the coverage. Although SentiStrength has presented good Macro-F1 values, its coverage is usually low as this method tends to classify a high number of instances as neutral. Note that some datasets provided by the SentiStrength's authors, as shown in Table~\ref{tab:labeled_dataset}, specially the Twitter datasets, have more neutral sentences than positive and negative ones. Another expected result is the good Macro-F1 values obtained by Emoticons, specially in the Twitter datasets. It is important to highlight  that, in spite of achieving high accuracy and Macro-F1, the coverage of many methods, such as PANAS, VADER, and SentiStrength, is low (e.g. below 30\%) as they only infer the polarity of part of the input sentences.Thus, the choice of a sentiment analysis is highly dependent on the data and application, suggesting that researchers and practitioners need to take into account  this tradeoff between prediction performance and coverage. 

The same high variability regarding the methods's prediction performance can be noted for the 3-class experiments, as presented in Table~\ref{tab:tabelao}. Umigon, the best method in five Twitter datasets, felt to the eighteenth place in the Comments\_NYT dataset. We can also note the lower Macro-F1 values for some methods like Emoticons are due to the high number of sentences without emoticons in the datasets. Methods like Emoticons DS and PANAS tend do classify only a small part of instances as neutral and also presented a poor performance in the 3-class experiments. Methods like SenticNet and LIWC were not originally developed for detecting neutral sentences and also achieved low values of Macro-F1. However, they also do not appear among the best methods in the 2-class experiments, which is the task they were originally designed for. This observation about LIWC is not valid for the newest version, as LIWC15 appears among the top five methods for 2-class and 3-class experiments (see Table~\ref{tab:meanrank}).

\begin{table*}[htpb]
  \centering
 \footnotesize
  \caption{Mean Rank Table for All Datasets}
    \begin{tabular}{|c|l|c|c|l|c|c|}
    \hline
    \multicolumn{3}{|c|}{3-Classes} & \multicolumn{4}{c|}{2-Classes}\\
    \hline
    \textbf{Pos} & \textbf{Method} & \textbf{Mean Rank} & \textbf{Pos} & \textbf{Method} & \textbf{Mean Rank} & \textbf{Coverage (\%)} \\ \hline
1 & VADER & 4.00(4.17)  &  1 & SentiStrength & 2.33(3.00) & 29.30(28.91) \\ \hline
2 & LIWC15 & 4.62  &  2 & Sentiment140 & 3.44 & 39.29  \\ \hline
3 & AFINN & 4.69  &  3 & Semantria & 4.61 & 62.34  \\ \hline
4 & Opinion Lexicon & 5.00  &  4 & Opinion Lexicon & 6.72 & 69.50  \\ \hline
5 & Semantria & 5.31  &  5 & LIWC15 & 7.33 & 68.28  \\ \hline
6 & Umigon & 5.77  &  6 & SO-CAL & 7.61 & 72.64  \\ \hline
7 & SO-CAL & 7.23  &  7 & AFINN & 8.11 & 73.05  \\ \hline
8 & Pattern.en & 9.92  &  8 & VADER & 9.17(9.79) & 82.20(83.18)  \\ \hline
9 & Sentiment140 & 10.92  &  9 & Umigon & 9.39 & 64.11  \\ \hline
10 & Emolex & 11.38  &  10 & PANAS-t & 10.17 & 5.10  \\ \hline
11 & Opinion Finder & 13.08  &  11 & Emoticons & 10.39 & 10.69  \\ \hline
12 & SentiWordNet & 13.38  &  12 & Pattern.en & 12.61 & 65.02  \\ \hline
13 & Sentiment140\_L & 13.54  &  13 & SenticNet & 13.61 & 84.00  \\ \hline
14 & SenticNet & 13.62  &  14 & Emolex & 14.50 & 66.12  \\ \hline
15 & SentiStrength & 13.69(13.71)  &  15 & Opinion Finder & 14.72 & 46.63  \\ \hline
16 & SASA & 14.77  &  16 & USent & 14.89 & 44.00  \\ \hline
17 & Stanford DM & 15.85  &  17 & Sentiment140\_L & 14.94 & 93.36  \\ \hline
18 & USent & 15.92  &  18 & NRC Hashtag & 17.17 & 93.52  \\ \hline
19 & NRC Hashtag & 16.31  &  19 & Stanford DM & 17.39 & 87.32  \\ \hline
20 & LIWC & 16.46  &  20 & SentiWordNet & 17.50 & 61.77  \\ \hline
21 & ANEW\_SUB & 18.54  &  21 & SASA & 18.94 & 60.12  \\ \hline
22 & Emoticons & 21.00  &  22 & LIWC & 19.67 & 61.82  \\ \hline
23 & PANAS-t & 21.77  &  23 & ANEW\_SUB & 21.17 & 94.20  \\ \hline
24 & Emoticons DS & 23.23  &  24 & Emoticons DS & 23.61 & 99.36  \\ \hline

    \end{tabular}%
  \label{tab:meanrank}%
\end{table*}%

Finally, Table~\ref{tab:friedman} presents the Friedman's test results showing that there are significant differences in the mean rankings observed for the methods across all datasets. It statistically indicates that in terms of accuracy and Macro-F1 there is no single method that always achieves a consistent rank position for different datasets, which is something similar to the well-known ``no-free lunch theorem''~\cite{Wolpert97nofree}. So, overall, before using a sentiment analysis method in a novel dataset, it is crucial to test different methods in a sample of data before simply choose one that is acceptable by the research community.

\begin{table}[h!]\centering
  \caption{Friedman's Test Results}
\begin{tabular}{ | l | l | l | l | }
\hline
	\multicolumn{2}{|c|}{2-class experiments} & \multicolumn{2}{c|}{3-class experiments} \\ \hline
	FR & 275.59 & FR & 197.52 \\ \hline
	Critical Value & 35.17 & Critical Value & 35.17 \\ \hline
	\multicolumn{2}{|c|}{\textbf{Reject null hypothesis}} & \multicolumn{2}{c|}{\textbf{Reject null hypothesis}}  \\ \hline
\end{tabular}
  \label{tab:friedman}
\end{table}

This last results suggests that, even with the good insights provided by this work about which methods perform better in each context, a preliminary investigation needs to be performed when sentiment analysis is used in a new dataset in order to guarantee a reasonable prediction performance. In the case in which prior tests are not feasible, this benchmark presents valuable information for researchers and companies that are planning to develop research and solutions on sentiment analysis.

\noindent
\textbf{Existing methods let space for improvements}:
We can note that the performance of the evaluated methods are ok, but there is a lot of space for improvements. For example, if we look at the Macro-F1 values only for the best method on each dataset (See Table~\ref{tab:tabelao2} and Table~\ref{tab:tabelao}), we can note that the overall prediction performance of the methods is still low -- i.e. Macro-F1 values are around 0.9 only for methods with low coverage in the 2-class experiments and only 0.6  for the 3-class experiment. Considering that we are looking at the performance of the best methods out of 24 unsupervised tools, these numbers suggest that current sentence-level sentiment analysis methods still let a lot of space for improvements. Additionally, we also noted that the best method for each dataset varies considerably from one dataset to another. This  might indicate that each method complements
the others in different ways.

\begin{figure*}[ht]
\centering
\includegraphics[width=18cm,height=\textheight,keepaspectratio]{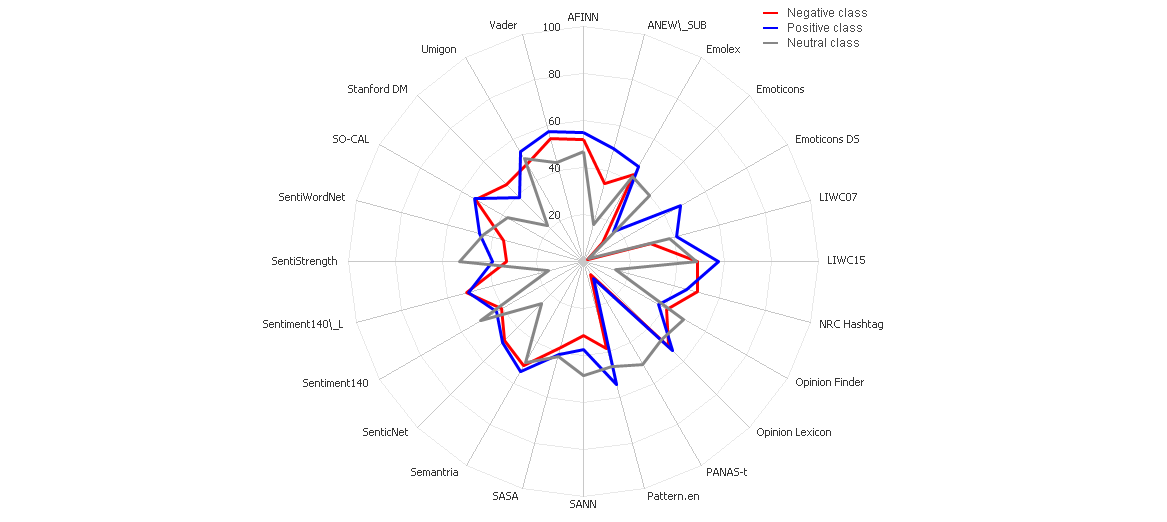}
\caption{Average F1 Score for each class.}
\label{fig:f1-class}
\end{figure*}

\noindent
\textbf{Most methods are better to classify positive than negative or neutral sentences}:
Figure~\ref{fig:f1-class} presents the average  F1 Score for the 3-class experiments. It is easier to notice that twelve out of twenty-four methods are more accurate while classifying positive than negative or neutral messages, suggesting that some methods may be more biased towards positivity. Neutral messages showed to be even harder to detect by most methods. 

Interestingly, recent efforts show that human language have a universal positivity bias (\cite{Garcia2012} and \cite{Dodds24022015}). Naturally, part of the bias is observed in sentiment prediction, an intrinsic property of some methods due to the way they are designed. For instance, \cite{hannak-2012-weather} developed a lexicon  in which positive and negative values are associated to words, hashtags, and any sort of tokens according to the frequency with which  these tokens appear in tweets containing positive and negative emoticons. This method showed to be biased towards positivity due to the larger amount of positivity in the data they used to build the lexicon. The overall poor performance of this specific method is credited to its lack of treatment of neutral messages and the focus on Twitter messages.

\noindent
\textbf{Some methods are consistently among the best ones}:
\if 0 Table~\ref{tab:meanrank} presents the mean rank value, detailed before, for 2-class and 3-class experiments. The elements are sorted by the overall mean rank each method achieved based on Macro-F1 for all datasets.  \red{The top seven methods based on Macro-F1 are SentiStrength, Semantria, AFINN, OpinionLexicon, Umigon, VADER and SO-CAL. This means that these methods produce the best results  across several datasets for both, 2 and 3-class tasks.}  These methods would be preferable in situations in which any sort of preliminary evaluation is not possible to be done. The mean rank for 2-class experiments is accompanied by the coverage metric,  which is very important to avoid misinterpretation of the results. Observe that SentiStrength exhibited the best mean rank for these  experiments, however it presents very low coverage, around 30\%,  a very poor result compared with Semantria and OpinionLexicon that achieved a worse mean rank (3.61 and 6.28, respectively) but an expressive better coverage, above 60\%.\fi

Table~\ref{tab:meanrank} presents the mean rank value, detailed before, for 2-class and 3-class experiments. The elements are sorted by the overall mean rank each method achieved based on Macro-F1 for all datasets.  The top nine methods based on Macro-F1 for the 2-class experiments are: SentiStrength, Sentiment140, Semantria, OpinionLexicon, LIWC15, SO-CAL, AFINN and VADER and Umigon. With the exception of Sentistrength,  replaced by Pattern.en, the other eight methods produce the best results  across several datasets for both, 2 and 3-class tasks.  These methods would be preferable in situations in which any sort of preliminary evaluation is not possible to be done. The mean rank for 2-class experiments is accompanied by the coverage metric,  which is very important to avoid misinterpretation of the results. Observe that SentiStrength and Sentiment140 exhibited the best mean ranks for these  experiments, however both present very low coverage, around 30\% and 40\%,  a very poor result compared with Semantria and OpinionLexicon that achieved a worse mean rank (4,61 and 6,62 respectively) but an expressive better coverage, above 60\%.  Note also that SentiStrength and Sentiment140 present poor results in the 3-class experiments which can be explained by their bias to the neutral class as mentioned before.

Another interesting finding is the fact that  VADER, the best method in the 3-class experiments, did not achieve the first position for none of the datasets. It reachs the second place five times, the third place twice, the seventh three times, and the fourth, sixth and fifth just once. It was a special case of consistency across all datasets. Tables \ref{tab:best_2class} and \ref{tab:best_2class} present the best method for each dataset in the 2-class and 3-class experiments, respectively.

\begin{table*}[htpb]
	\centering
	\caption{Best Method for each Dataset - 2-class experiments}
\begin{tabular}{ | l | l | l | l | l | l | }\hline
\textbf{Dataset} & \textbf{Method} & \textbf{F1-Pos} & \textbf{F1-Neg} & \textbf{Macro-F1} & \textbf{Coverage}
 \\ \hline

	Comments\_BBC & SentiStrength & 70.59 & 96.61 & 83.60 & 32.85 \\ \hline
	Comments\_Digg & SentiStrength & 84.96 & 94.64 & 89.80 & 27.49 \\ \hline
	Comments\_NYT & SentiStrength & 70.11 & 86.52 & 78.32 & 17.63 \\ \hline
	Comments\_TED & Emoticons & 85.71 & 94.12 & 89.92 & 1.65 \\ \hline
	Comments\_YTB & SentiStrength & 96.94 & 89.62 & 93.28 & 38.24 \\ \hline
	Reviews\_I & SenticNet & 97.39 & 93.66 & 95.52 & 69.41 \\ \hline
	Reviews\_II & SenticNet & 94.15 & 93.87 & 94.01 & 94.25 \\ \hline
	Myspace & SentiStrength & 98.73 & 88.46 & 93.6 & 31.53 \\ \hline
	Amazon & SentiStrength & 93.85 & 79.38 & 86.62 & 19.58 \\ \hline
	Tweets\_DBT & Sentiment140 & 72.86 & 83.55 & 78.2 & 18.75 \\ \hline
	Tweets\_RND\_I & SentiStrength & 95.28 & 90.6 & 92.94 & 27.13 \\ \hline
	Tweets\_RND\_II & VADER & 99.31 & 98.45 & 98.88 & 94.4 \\ \hline
	Tweets\_RND\_III & Sentiment140 & 97.57 & 95.9 & 96.73 & 50.77 \\ \hline
	Tweets\_RND\_IV & Emoticons & 94.74 & 86.76 & 88.6 & 58.27 \\ \hline
	Tweets\_STF & SentiStrength & 95.76 & 94.81 & 95.29 & 41.78 \\ \hline
	Tweets\_SAN & SentiStrength & 90.23 & 88.59 & 89.41 & 29.61 \\ \hline
	Tweets\_Semeval & SentiStrength & 93.93 & 83.4 & 88.66 & 28.66 \\ \hline
	RW & SentiStrength & 90.04 & 75.79 & 82.92 & 23.12 \\ \hline

\end{tabular}
	\label{tab:best_2class}
\end{table*}

\begin{table*}[htpb]
	\centering
	\caption{Best Method for each Dataset - 3-class experiments}
\begin{tabular}{ | l | l | l | l | l | l | }\hline
\textbf{Dataset} & \textbf{Method} & \textbf{F1-Pos} & \textbf{F1-Neg} & \textbf{F1-Neu} & \textbf{Macro-F1}
 \\ \hline

	Comments\_BBC & Semantria & 36.76 & 67.94 & 43.49 & 49.40 \\ \hline
	Comments\_Digg & Umigon & 49.62 & 62.04 & 44.27 & 51.98 \\ \hline
	Comments\_NYT & SO-CAL & 56.99 & 60.08 & 15.34 & 44.14 \\ \hline
	Comments\_TED & Opinion Lexicon & 64.95 & 56.59 & 30.77 & 50.77 \\ \hline
	Comments\_YTB & LIWC15 & 73.68 & 49.72 & 48.79 & 57.4 \\ \hline
	Myspace & LIWC15 &78.83 & 41.74 & 43.76 & 54.78\\ \hline
	Tweets\_DBT & Opinion Lexicon & 43.44 & 47.71 & 48.84 & 46.66 \\ \hline
	Tweets\_RND\_I & Umigon & 60.53 & 51.39 & 65.22 & 59.05 \\ \hline
	Tweets\_RND\_III & Umigon & 63.33 & 57.00 & 82.10 & 67.47 \\ \hline
	Tweets\_RND\_IV & Umigon & 75.86 & 76.33 & 71.54 & 74.58 \\ \hline
	Tweets\_SAN & Umigon & 44.16 & 45.95 & 70.45 & 53.52 \\ \hline
	Tweets\_Semeval & Umigon & 64.28 & 46.41 & 73.13 & 61.27 \\ \hline
	RW & Sentiment140 & 62.24 & 51.17 & 42.66 & 52.02 \\ \hline

\end{tabular}
	\label{tab:best_3class}
\end{table*}

\if 0 falar do VADER
posições para três classes --> 5,2,2,2,2,2,3,3,7,7,6,4,7,
\fi

\if 0
\green{These results proves that many negative and positive sentences are classified as neutral by SentiStrength and Sentiment140}.
\fi

\noindent
\textbf{Methods are often better in the datasets they were originally evaluated}:
We also note those methods perform better in datasets in which they were originally validated, which is somewhat expected due to fine tuning procedures. We could do this comparison only for SentiStrength and VADER, which kindly allowed the entire reproducibility of their work, sharing both methods and datasets.
To understand this difference,  we calculated the mean rank for these methods without their `original' datasets and put the results in parenthesis. Note that, in some cases the rank order changes towards a lower value but it does not imply in major changes.
We also note those methods often perform better in datasets in which they were originally validated, which is somewhat expected due to fine tuning procedures. We could do this comparison only for SentiStrength and VADER, which kindly allowed the entire reproducibility of their work, sharing both methods and datasets.
To understand this difference,  we calculated the mean rank for these methods without their `original' datasets and put the results in parenthesis. Note that, in some cases the rank order slightly changes but it does not imply in major changes.
Overall, these observations suggest that initiatives like SemEval are key for the development of the area, as they allow methods to compete in a contest for a specific dataset. More important, it highlight that a standard sentiment analysis benchmark is needed and it needs to be constantly updated.
We also emphasize that is possible that other methods, such as paid softwares, make use of some of the datasets used in this benchmark to improve their performance as most of gold standard used in this work is available in the Web or under request to authors.
%Since we cannot confirm if this indeed  has occurred we cannot produce mean rank variations explicitly for the other methods as we did for SentiStrength and VADER.

\noindent
\textbf{Some methods showed to be better for specific contexts}:
In order to better understand the prediction performance of methods in types of data, we divided all datasets in three specific contexts --  Social Networks, Comments, and Reviews -- and calculated mean rank of the methods for each of them. Table~\ref{tab:contexts} presents the contexts and the respective datasets.

\begin{table}[h!]
	\centering
  \caption{Contexts' Groups}
\begin{tabular}{ | l | M{5.4cm} | }
\hline
	\multicolumn{2}{|c|}{Context Groups}  \\ \hline
	Social Networks & Myspace, Tweets\_DBT, Tweets\_RND\_I, Tweets\_RND\_II, Tweets\_RND\_III, Tweets\_RND\_IV, Tweets\_STF, Tweets\_SAN, Tweets\_Semeval \\ \hline
	Comments & Comments\_BBC, Comments\_DIGG, Comments\_NYT, Comments\_TED, Comments\_YTB, RW \\ \hline
	Reviews & Reviews\_I, Reviews\_I, Amazon \\ \hline
\end{tabular}
  \label{tab:contexts}
\end{table}

Tables~\ref{tab:meanrank_social}, \ref{tab:meanrank_comments} and~\ref{tab:meanrank_reviews}  present the mean rank for each context separately. In the context of Social Networks the best method for 3-class experiments was Umigon, followed by LIWC15 and VADER. In the case of 2-class the winner was SentiStrength with a coverage around 30\%  and the third and sixth place were Emoticons and Panas-t with about 18\% and 6\% of coverage, respectively. This highlights the importance to analyze the 2-class results together with the coverage. Overall, when there is an emoticon on the text or a word from the psychometric scale PANAS, these methods are able to tell the polarity of the sentences, but they are not able to identify the polarity of the input text for the large majority of the input text. Recent efforts suggest these properties are useful for combination of methods~\cite{Pollyanna@COSN13}. Sentiment140, LIWC15, Semantria, OpinionLexicon and Umigon showed to be the best alternatives for detecting only positive and negative polarities in social network data due to the high coverage and prediction performance. It is important to highlight that LIWC 2007 appears on the 16th and 21th position for the 3-class and 2-class mean rank results for the social network datasets and it is a very popular method in this community. On the other side, the newest version of LIWC (2015) presented a considerable evolution obtaining the second and the fourth place in the same datasets.

\begin{table*}[htpb]
  \centering
  \footnotesize
  \caption{Mean Rank Table for Datasets of Social Networks}
    \begin{tabular}{|c|l|c|c|l|c|c|}
    \hline
    \multicolumn{3}{|c|}{3-Classes} & \multicolumn{4}{c|}{2-Classes}\\
    \hline
    \textbf{Pos} & \textbf{Method} & \textbf{Mean Rank} & \textbf{Pos} & \textbf{Method} & \textbf{Mean Rank} & 
    \textbf{Coverage (\%)}\\ \hline

1 & Umigon & 2.57  &  1 & SentiStrength & 2.22(2.57) & 31.54(32.18) \\ \hline
2 & LIWC15 & 3.29  &  2 & Sentiment140 & 3.00 & 46.98  \\ \hline
3 & VADER & 4.57(4.57)  &  3 & Emoticons & 5.11 & 18.04  \\ \hline
4 & AFINN & 5.00  &  4 & LIWC15 & 5.67 & 71.73  \\ \hline
5 & Opinion Lexicon & 5.57  &  5 & Semantria & 5.89 & 61.98  \\ \hline
6 & Semantria & 6.00  &  6 & PANAS-t & 6.33 & 5.87  \\ \hline
7 & Sentiment140 & 7.00  &  7 & Opinion Lexicon & 7.56 & 66.56  \\ \hline
8 & Pattern.en & 7.57  &  8 & Umigon & 8.00 & 71.67  \\ \hline
9 & SO-CAL & 9.00  &  9 & AFINN & 8.67 & 73.37  \\ \hline
10 & Emolex & 12.29  &  10 & SO-CAL & 8.78 & 67.81  \\ \hline
11 & SentiStrength & 12.43(11.60)  &  11 & VADER & 8.78(9.75) & 83.29(81.90)  \\ \hline
12 & Opinion Finder & 13.00  &  12 & Pattern.en & 11.22 & 69.47  \\ \hline
13 & SentiWordNet & 13.57  &  13 & Sentiment140\_L & 14.00 & 94.61  \\ \hline
14 & SenticNet & 14.14  &  14 & Opinion Finder & 14.33 & 39.58  \\ \hline
15 & SASA & 14.86  &  15 & Emolex & 14.56 & 62.63  \\ \hline
16 & LIWC & 15.43  &  16 & USent & 15.22 & 38.60  \\ \hline
17 & Sentiment140\_L & 15.43  &  17 & SenticNet & 17.22 & 75.46  \\ \hline
18 & USent & 16.00  &  18 & SentiWordNet & 18.44 & 61.41  \\ \hline
19 & ANEW\_SUB & 19.14  &  19 & NRC Hashtag & 19.11 & 94.20  \\ \hline
20 & Emoticons & 19.14  &  20 & SASA & 19.44 & 58.57  \\ \hline
21 & Stanford DM & 19.43  &  21 & LIWC & 19.56 & 61.24  \\ \hline
22 & NRC Hashtag & 20.00  &  22 & ANEW\_SUB & 20.56 & 93.51  \\ \hline
23 & PANAS-t & 20.86  &  23 & Stanford DM & 22.56 & 89.06  \\ \hline
24 & Emoticons DS & 23.71  &  24 & Emoticons DS & 23.78 & 99.28  \\ \hline
    \end{tabular}%
  \label{tab:meanrank_social}%
\end{table*}%

\begin{table*}[htbp]
  \centering
  \footnotesize
  \caption{Mean Rank Table for Datasets of Comments}
    \begin{tabular}{|c|l|c|c|l|c|c|}
    \hline
    \multicolumn{3}{|c|}{3-Classes} & \multicolumn{4}{c|}{2-Classes}\\
    \hline
    \textbf{Pos} & \textbf{Method} & \textbf{Mean Rank} & \textbf{Pos} & \textbf{Method} & \textbf{Mean Rank}& \textbf{Coverage (\%)}\\ \hline

1 & VADER & 3.33(3.60)  &  1 & SentiStrength & 1.17(1.50) & 28.29(24.02) \\ \hline
2 & AFINN & 4.33  &  2 & Semantria & 2.83 & 61.02  \\ \hline
3 & Opinion Lexicon & 4.33  &  3 & Sentiment140 & 4.17 & 36.49  \\ \hline
4 & Semantria & 4.50  &  4 & Opinion Lexicon & 6.50 & 71.59  \\ \hline
5 & SO-CAL & 5.17  &  5 & LIWC15 & 6.67 & 65.80  \\ \hline
6 & LIWC15 & 6.17  &  6 & AFINN & 7.00 & 74.21  \\ \hline
7 & Umigon & 9.50  &  7 & SO-CAL & 7.50 & 74.59  \\ \hline
8 & Emolex & 10.33  &  8 & VADER & 9.50(9.60) & 81.98(85.34)  \\ \hline
9 & Sentiment140\_L & 11.33  &  9 & Umigon & 10.50 & 57.87  \\ \hline
10 & Stanford DM & 11.67  &  10 & Emoticons & 11.83 & 4.99  \\ \hline
11 & NRC Hashtag & 12.00  &  11 & Opinion Finder & 13.00 & 55.66  \\ \hline
12 & Pattern.en & 12.67  &  12 & SenticNet & 13.00 & 95.28  \\ \hline
13 & SenticNet & 13.00  &  13 & USent & 14.00 & 45.66  \\ \hline
14 & Opinion Finder & 13.17  &  14 & NRC Hashtag & 14.67 & 93.43  \\ \hline
15 & SentiWordNet & 13.17  &  15 & Emolex & 15.00 & 69.69  \\ \hline
16 & SASA & 14.67  &  16 & PANAS-t & 15.50 & 5.10  \\ \hline
17 & SentiStrength & 15.17(19.00)  &  17 & Stanford DM & 15.67 & 84.43  \\ \hline
18 & Sentiment140 & 15.50  &  18 & Pattern.en & 15.83 & 59.00  \\ \hline
19 & USent & 15.83  &  19 & Sentiment140\_L & 15.83 & 92.30  \\ \hline
20 & LIWC & 17.67  &  20 & SentiWordNet & 17.00 & 63.32  \\ \hline
21 & ANEW\_SUB & 17.83  &  21 & SASA & 17.50 & 61.91  \\ \hline
22 & Emoticons DS & 22.67  &  22 & LIWC & 19.67 & 62.24  \\ \hline
23 & PANAS-t & 22.83  &  23 & ANEW\_SUB & 22.00 & 94.31  \\ \hline
24 & Emoticons & 23.17  &  24 & Emoticons DS & 23.67 & 99.31  \\ \hline
    \end{tabular}%
  \label{tab:meanrank_comments}%
\end{table*}%

\begin{table*}[htbp]
  \centering
  \footnotesize
  \caption{Mean Rank Table for Datasets of Reviews}
    \begin{tabular}{|c|l|c|c|l|c|c|}
    \hline
    \multicolumn{3}{|c|}{3-Classes} & \multicolumn{4}{c|}{2-Classes}\\
    \hline
    \textbf{Pos} & \textbf{Method} & \textbf{Mean Rank} & \textbf{Pos} & \textbf{Method} & \textbf{Mean Rank}& \textbf{Coverage (\%)}\\ \hline

1 & - & -  &  1 & Sentiment140 & 3.33 & 21.82  \\ \hline
2 & - & -  &  2 & SenticNet & 4.00 & 87.05  \\ \hline
3 & - & -  &  3 & Semantria & 4.33 & 66.04  \\ \hline
4 & -  & -  &  4 & SO-CAL & 4.33 & 83.20  \\ \hline
5 & - & -  &  5 & Opinion Lexicon & 4.67 & 74.14  \\ \hline
6 & - & -  &  6 & SentiStrength & 5.00(5.00) & 24.56(24.56) \\ \hline
7 & - & -  &  7 & Stanford DM & 5.33 & 87.89  \\ \hline
8 & - & -  &  8 & AFINN & 8.67 & 69.77  \\ \hline
9 & - & -  &  9 & VADER & 9.67(11.00) & 79.39(82.70)  \\ \hline
10 & - & -  &  10 & Pattern.en & 10.33 & 63.70  \\ \hline
11 & - & -  &  11 & PANAS-t & 11.00 & 2.80  \\ \hline
12 & - & -  &  12 & Umigon & 11.33 & 53.90  \\ \hline
13 & - &-  &  13 & Emolex & 13.33 & 69.47  \\ \hline
14 & - & -  &  14 & LIWC15 & 13.67 & 62.90  \\ \hline
15 & - & -  &  15 & USent & 15.67 & 56.85  \\ \hline
16 & - & -  &  16 & SentiWordNet & 15.67 & 59.73  \\ \hline
17 & - & -  &  17 & Sentiment140\_L & 16.00 & 91.71  \\ \hline
18 & - & -  &  18 & NRC Hashtag & 16.33 & 91.64  \\ \hline
19 & - & -  &  19 & Opinion Finder & 19.33 & 49.73  \\ \hline
20 & - & - &  20 & LIWC & 20.00 & 62.75  \\ \hline
21 & - & -  &  21 & SASA & 20.33 & 61.22  \\ \hline
22 & - & -  &  22 & ANEW\_SUB & 21.33 & 96.05  \\ \hline
23 & - & -  &  23 & Emoticons DS & 23.00 & 99.71  \\ \hline
24 & -  & -  &  24 & Emoticons & 23.33 & 0.04  \\ \hline

    \end{tabular}%
  \label{tab:meanrank_reviews}%
\end{table*}%

Similar analyses can be performed for the contexts Comments and Reviews. Sentistregth, VADER, Semantria, AFINN, and Opinion Lexicon showed to be the best alternatives for 2-class and 3-class experiments on datasets of comments whereas Sentiment140, SenticNet, Semantria and SO-CAL showed to be the best for the 2-class experiments for the datasets containing short reviews. Note that for the last one, the 3-class experiments have no results since datasets containing  reviews have no neutral sentences nor a representative number of sentences without subjectivity.

We also calculated the Friedman's value for each of these specific contexts. Even after grouping the datasets, we still observe that there are significant differences in the observed ranks across the datasets. Although the values obtained for each context were quite smaller than Friedman' global value, they are still above the critical value. Table~\ref{tab:friedman_contexts} presents the results of Friedman's test for the individual contexts in both experiments, 2 and 3-class. Recall that for the 3-class experiments, datasets with no neutral sentences or with an unrepresentative number of neutral sentences were not considered. For this reason, Friedman's results for 3-class experiments in the Reviews context presents no values.

\begin{table}[h!]\centering
  \caption{Friedman's Test Results per Contexts}
\begin{tabular}{ | l | l | l | l | }
\hline
	\multicolumn{4}{|c|}{Context: Social Networks} \\ \hline
	\multicolumn{2}{|c|}{2-class experiments} & \multicolumn{2}{c|}{3-class experiments} \\ \hline
	FR & 175.94 & FR & 124.16 \\ \hline
	Critical Value & 35.17 & Critical Value & 35.17 \\ \hline
	\multicolumn{2}{|c|}{\textbf{Reject null hypothesis}} & \multicolumn{2}{c|}{\textbf{Reject null hypothesis}}  \\ \hline
	\multicolumn{4}{|c|}{Context: Comments} \\ \hline
	\multicolumn{2}{|c|}{2-class experiments} & \multicolumn{2}{c|}{3-class experiments} \\ \hline
	FR & 95.59 & FR & 96.41 \\ \hline
	Critical Value & 35.17 & Critical Value & 35.17 \\ \hline
	\multicolumn{2}{|c|}{\textbf{Reject null hypothesis}} & \multicolumn{2}{c|}{\textbf{Reject null hypothesis}}  \\ \hline
	\multicolumn{4}{|c|}{Context: Reviews} \\ \hline
	\multicolumn{2}{|c|}{2-class experiments} & \multicolumn{2}{c|}{3-class experiments} \\ \hline
	FR & 60.52 & FR & - \\ \hline
	Critical Value & 35.17 & Critical Value & - \\ \hline
	\multicolumn{2}{|c|}{\textbf{Reject null hypothesis}} & \multicolumn{2}{c|}{\textbf{Reject null hypothesis}}  \\ \hline			
\end{tabular}
  \label{tab:friedman_contexts}
\end{table}

\section*{Concluding Remarks}

Recent efforts to analyze the moods embedded in Web 2.0 content have adopted various sentiment analysis methods, which were originally developed in linguistics and psychology. Several of these methods became widely used in their knowledge fields and have now been applied as tools to quantify moods in the context of unstructured short messages in online social networks. In this article, we present a thorough comparison of twenty-four popular sentence-level sentiment analysis methods using gold standard datasets that span different types of data sources. Our effort quantifies the prediction performance of the twenty-four popular sentiment analysis methods across eighteen datasets for two tasks: differentiating two classes (positive and negative) and three classes (positive, negative, and neutral). 
 
Among many findings, we highlight that although our results identified a few methods  able to appear among the best ones for different datasets, we noted that the overall prediction performance still left a lot of space for improvements. More important, we show that the prediction performance of methods vary  largely across datasets. For example, LIWC 2007, is among the most popular sentiment methods in the social network context and obtained a bad rank position in comparison with other datasets. This suggests that sentiment analysis methods cannot be used as ``off-the-shelf'' methods, specially for novel datasets. We show that the same social media text can be interpreted very differently depending on the choice of a sentiment method, suggesting that it is important that researchers and companies perform experiments with different methods before applying a method.

%We also show that sentiment analysis methods are biased towards positivity, which is both a limitation of these methods and can have implications for specially those built based on human usage of the language.

%In this article we believe we presented a relevant academic contribution to the research area of sentiment analysis. We produced detailed results regarding the performance of a considerable number methods in a large number of datasets covering several different contexts. Our results can be quite useful specially for state-of-the-practice usage when there is no previous effort on detecting the methods that may fit a research our practical need. We are also releasing a Web system so that  others  can easily compare results of those methods in their own datasets. With this system one could easily test which method would be most suitable for a particular dataset and application. We hope that our tool will not only help researchers and practitioners to compare a wide range of sentiment analysis techniques but also that it can help foster new relevant research in this area with a rigorous scientific approach.

As a final contribution  we open the datasets and codes used in this paper for the research community. We also incorporated them in a Web service from our research team called iFeel~\cite{araujo2016@icwsm} \if 0 \footnote{http://www.ifeel.dcc.ufmg.br} \fi that allow users to easily compare the results of various sentiment analysis methods. We hope our effort can not only help researchers and practitioners to compare a wide range of sentiment analysis techniques, but also help fostering new relevant research in this area with a rigorous scientific approach.

%\green{as a benchmark just like CPU benchmarks are used to evaluate efficiency of processors}.

%While our effort is first of a kind, there still remain popular sentiment analysis methods that are yet to be validated such as the Profile of Mood States (POMS)~\cite{Bollen}. We hope to be able to implement these efforts and add them to our comparisons in the future.

\bibliographystyle{SIGCHI-Reference-Format}
\bibliography{references}


\begin{thebibliography}{00}

%%% the Web address.

\ifx \showCODEN    \undefined \def \showCODEN     #1{\unskip}     \fi
\ifx \showDOI      \undefined \def \showDOI       #1{{\tt DOI:}\penalty0{#1}\ }
  \fi
\ifx \showISBNx    \undefined \def \showISBNx     #1{\unskip}     \fi
\ifx \showISBNxiii \undefined \def \showISBNxiii  #1{\unskip}     \fi
\ifx \showISSN     \undefined \def \showISSN      #1{\unskip}     \fi
\ifx \showLCCN     \undefined \def \showLCCN      #1{\unskip}     \fi
\ifx \shownote     \undefined \def \shownote      #1{#1}          \fi
\ifx \showarticletitle \undefined \def \showarticletitle #1{#1}   \fi
\ifx \showURL      \undefined \def \showURL       #1{#1}          \fi

\bibitem{ABBASI14.483}
{Ahmed Abbasi}, {Ammar Hassan}, {and} {Milan Dhar}. 2014.
\newblock \showarticletitle{Benchmarking Twitter Sentiment Analysis Tools}. In
  {\em Proceedings of the Ninth International Conference on Language Resources
  and Evaluation (LREC'14)} (26-31), {Nicoletta Calzolari~(Conference Chair)},
  {Khalid Choukri}, {Thierry Declerck}, {Hrafn Loftsson}, {Bente Maegaard},
  {Joseph Mariani}, {Asuncion Moreno}, {Jan Odijk}, {and} {Stelios Piperidis}
  (Eds.). European Language Resources Association (ELRA), Reykjavik, Iceland.
\newblock
\showISBNx{978-2-9517408-8-4}


\bibitem{aisopos2014}
{Fotis Aisopos}. 2014.
\newblock Manually Annotated Sentiment Analysis Twitter Dataset NTUA.
\newblock   (2014).
\newblock
\newblock
\shownote{\url{www.grid.ece.ntua.gr}.}


\bibitem{anew_wkb}
{Marc~Brysbaert Amy Beth~Warriner, Victor~Kuperman}. 2013.
\newblock \showarticletitle{Norms of valence, arousal, and dominance for 13,915
  English lemmas}.
\newblock {\em Behavior research methods\/} {45}, 4 (Dec. 2013), 1191--1207.
\newblock


\bibitem{araujo2016@icwsm}
{Matheus Araujo}, {João~P. Diniz}, {Lucas Bastos}, {Elias Soares}, {Manoel
  Júnior}, {Miller Ferreira}, {Filipe Ribeiro}, {and} {Fabr\'icio Benevenuto}.
  2016.
\newblock \showarticletitle{iFeel 2.0: A Multilingual Benchmarking System for
  Sentence-Level Sentiment Analysis}. In {\em Proceedings of the International
  AAAI Conference on Web-Blogs and Social Media}. Cologne, Germany.
\newblock


\bibitem{sentiwordnet3}
{Stefano Baccianella}, {Andrea Esuli}, {and} {Fabrizio Sebastiani}. 2010.
\newblock \showarticletitle{SentiWordNet 3.0: An Enhanced Lexical Resource for
  Sentiment Analysis and Opinion Mining.}. In {\em LREC} (2010-06-02),
  {Nicoletta Calzolari}, {Khalid Choukri}, {Bente Maegaard}, {Joseph Mariani},
  {Jan Odijk}, {Stelios Piperidis}, {Mike Rosner}, {and} {Daniel Tapias}
  (Eds.).
\newblock
\showISBNx{2-9517408-6-7}


\bibitem{friedman2013}
{Mark~L. Berenson}, {David~M. Levine}, {and} {Kathryn~A. Szabat}. 2014.
\newblock {\em Basic Business Statistics - Concepts and Applications\/} (13
  ed.).
\newblock Pearson, USA. 840 pages.
\newblock
\showISBNx{978-0321946393}


\bibitem{Biever2010}
{Celeste Biever}. 2010.
\newblock \showarticletitle{Twitter mood maps reveal emotional states of
  America}.
\newblock {\em The New Scientist\/}  {207} (2010).
\newblock
Issue 2771.


\bibitem{DBLP:journals/corr/abs-1010-3003}
{Johan Bollen}, {Huina Mao}, {and} {Xiao-Jun Zeng}. 2010.
\newblock \showarticletitle{{Twitter Mood Predicts the Stock Market}}.
\newblock {\em CoRR\/}  {abs/1010.3003} (2010).
\newblock


\bibitem{Bollen}
{Johan Bollen}, {Alberto Pepe}, {and} {Huina Mao}. 2009.
\newblock \showarticletitle{{Modeling Public Mood and Emotion: Twitter
  Sentiment and Socio-Economic Phenomena}}.
\newblock {\em CoRR\/}  {abs/0911.1583} (2009).
\newblock


\bibitem{citeulike:3519108}
{M.~M. Bradley} {and} {P.~J. Lang}. 1999.
\newblock {\em {Affective norms for English words ({ANEW}): Stimuli,
  instruction manual, and affective ratings}}.
\newblock {T}echnical {R}eport. Center for Research in Psychophysiology,
  University of Florida, Gainesville, Florida.
\newblock


\bibitem{subtlex}
{Marc Brysbaert} {and} {Boris New}. 2009.
\newblock \showarticletitle{Moving beyond Kucera and Francis: A Critical
  Evaluation of Current Word Frequency Norms and the Introduction of a New and
  Improved Word Frequency Measure for American English}.
\newblock {\em Behavior research methods\/} {41}, 4 (2009), 977--990.
\newblock


\bibitem{senticnet3}
{E. Cambria}, {D. Olsher}, {and} {D. Rajagopal}. 2014.
\newblock \showarticletitle{Senticnet 3: A common and common-sense knowledge
  base for cognition-driven sentiment analysis}. In {\em AAAI}. Quebec City,
  1515--1521.
\newblock


\bibitem{FSS102216}
{Erik Cambria}, {Robert Speer}, {Catherine Havasi}, {and} {Amir Hussain}. 2010.
\newblock \showarticletitle{SenticNet: A Publicly Available Semantic Resource
  for Opinion Mining}. In {\em AAAI Fall Symposium Series}.
\newblock


\bibitem{cha_icwsm10}
{Meeyoung Cha}, {Hamed Haddadi}, {Fabricio Benevenuto}, {and} {Krishna~P.
  Gummadi}. 2010.
\newblock \showarticletitle{Measuring User Influence in Twitter: The Million
  Follower Fallacy}. In {\em International AAAI Conference on Weblogs and
  Social Media (ICWSM)}.
\newblock


\bibitem{de2012pattern}
{Tom De~Smedt} {and} {Walter Daelemans}. 2012.
\newblock \showarticletitle{Pattern for python}.
\newblock {\em The Journal of Machine Learning Research\/} {13}, 1 (2012),
  2063--2067.
\newblock


\bibitem{diakopoulos2010characterizing}
{N.A. Diakopoulos} {and} {D.A. Shamma}. 2010.
\newblock \showarticletitle{Characterizing debate performance via aggregated
  twitter sentiment}. In {\em Proceedings of the 28th international conference
  on Human factors in computing systems}. ACM, 1195--1198.
\newblock


\bibitem{Dodds24022015}
{Peter~Sheridan Dodds}, {Eric~M. Clark}, {Suma Desu}, {Morgan~R. Frank},
  {Andrew~J. Reagan}, {Jake~Ryland Williams}, {Lewis Mitchell}, {Kameron~Decker
  Harris}, {Isabel~M. Kloumann}, {James~P. Bagrow}, {Karine Megerdoomian},
  {Matthew~T. McMahon}, {Brian~F. Tivnan}, {and} {Christopher~M. Danforth}.
  2015.
\newblock \showarticletitle{Human language reveals a universal positivity
  bias}.
\newblock {\em Proceedings of the National Academy of Sciences\/} {112}, 8
  (2015), 2389--2394.
\newblock
\showDOI{%
\url{http://dx.doi.org/10.1073/pnas.1411678112}}


\bibitem{DoddsP2009Measuring}
{Peter~Sheridan Dodds} {and} {Christopher~M Danforth}. 2009.
\newblock \showarticletitle{Measuring the happiness of large-scale written
  expression: songs, blogs, and presidents}.
\newblock {\em Journal of Happiness Studies\/} {11}, 4 (2009), 441--456.
\newblock
\showDOI{%
\url{http://dx.doi.org/10.1007/s10902-009-9150-9}}


\bibitem{sentiwordnet}
{Esuli} {and} {Sebastiani}. 2006.
\newblock \showarticletitle{SentiWordNet: A Publicly Available Lexical Resource
  for Opinion Mining}. In {\em International Conference on Language Resources
  and Evaluation (LREC)}. 417--422.
\newblock


\bibitem{Feldman:2013:TAS:2436256.2436274}
{Ronen Feldman}. 2013.
\newblock \showarticletitle{Techniques and Applications for Sentiment
  Analysis}.
\newblock {\em Commun. ACM\/} {56}, 4 (April 2013), 82--89.
\newblock
\showISSN{0001-0782}
\showDOI{%
\url{http://dx.doi.org/10.1145/2436256.2436274}}


\bibitem{Garcia2012}
{David Garcia}, {Antonios Garas}, {and} {Frank Schweitzer}. 2012.
\newblock \showarticletitle{Positive words carry less information than negative
  words}.
\newblock {\em EPJ Data Science\/} {1}, 1 (2012), 1--12.
\newblock


\bibitem{Go2009}
{Alec Go}, {Richa Bhayani}, {and} {Lei Huang}. 2009.
\newblock \showarticletitle{Twitter Sentiment Classification using Distant
  Supervision}.
\newblock {\em Processing\/}  {-} (2009), 1--6.
\newblock


\bibitem{godbole2007large}
{Namrata Godbole}, {Manjunath Srinivasaiah}, {and} {Steven Skiena}. 2007.
\newblock \showarticletitle{Large-Scale Sentiment Analysis for News and Blogs}.
  In {\em Proceedings of the International Conference on Weblogs and Social
  Media (ICWSM)}.
\newblock


\bibitem{Pollyanna@COSN13}
{Pollyanna Gon\c{c}alves}, {Matheus Araujo}, {Fabrício Benevenuto}, {and}
  {Meeyoung Cha}. 2013a.
\newblock \showarticletitle{Comparing and Combining Sentiment Analysis
  Methods}. In {\em Proceedings of the 1st ACM Conference on Online Social
  Networks (COSN'13)}. 12.
\newblock


\bibitem{polly@panast}
{Pollyanna Gon\c{c}alves}, {Fabr\'{\i}cio Benevenuto}, {and} {Meeyoung Cha}.
  2013b.
\newblock \showarticletitle{{PANAS-t: A Pychometric Scale for Measuring
  Sentiments on Twitter}}.
\newblock   {abs/1308.1857v1} (2013).
\newblock


\bibitem{hannak-2012-weather}
{Aniko Hannak}, {Eric Anderson}, {Lisa~Feldman Barrett}, {Sune Lehmann}, {Alan
  Mislove}, {and} {Mirek Riedewald}. 2012.
\newblock \showarticletitle{Tweetin' in the Rain: Exploring societal-scale
  effects of weather on mood}. In {\em {Int'l AAAI Conference on Weblogs and
  Social Media (ICWSM)}}.
\newblock


\bibitem{Hu:2004:MSC:1014052.1014073}
{Minqing Hu} {and} {Bing Liu}. 2004.
\newblock \showarticletitle{Mining and summarizing customer reviews} {\em (KDD
  '04)}. 168--177.
\newblock
\showURL{%
\url{http://doi.acm.org/10.1145/1014052.1014073}}


\bibitem{hutto2014vader}
{CJ Hutto} {and} {Eric Gilbert}. 2014.
\newblock \showarticletitle{Vader: A parsimonious rule-based model for
  sentiment analysis of social media text}. In {\em Eighth International AAAI
  Conference on Weblogs and Social Media (ICWSM)}.
\newblock


\bibitem{jain91}
{Raj Jain}. 1991.
\newblock {\em The art of computer systems performance analysis - techniques
  for experimental design, measurement, simulation, and modeling.\/} (1 ed.).
\newblock Wiley, Canada. 1--685 pages.
\newblock
\showISBNx{978-0-471-50336-1}


\bibitem{Johnson015}
{Rie Johnson} {and} {Tong Zhang}. 2015.
\newblock \showarticletitle{Effective Use of Word Order for Text Categorization
  with Convolutional Neural Networks}. In {\em {NAACL} {HLT} 2015, The 2015
  Conference of the North American Chapter of the Association for Computational
  Linguistics: Human Language Technologies, Denver, Colorado, USA, May 31 -
  June 5, 2015}. 103--112.
\newblock


\bibitem{KalchbrennerACL2014}
{Nal Kalchbrenner}, {Edward Grefenstette}, {and} {Phil Blunsom}. 2014.
\newblock \showarticletitle{A Convolutional Neural Network for Modelling
  Sentences}. In {\em Proceedings of the 52nd Annual Meeting of the Association
  for Computational Linguistics}.
\newblock


\bibitem{kouloumpis2011twitter}
{Efthymios Kouloumpis}, {Theresa Wilson}, {and} {Johanna Moore}. 2011.
\newblock \showarticletitle{Twitter Sentiment Analysis: The Good the Bad and
  the OMG!}. In {\em {Int'l AAAI Conference on Weblogs and Social Media
  (ICWSM)}}.
\newblock


\bibitem{Kramer2014}
{Adam D~I Kramer}, {Jamie~E Guillory}, {and} {Jeffrey~T Hancock}. 2014.
\newblock \showarticletitle{{Experimental evidence of massive-scale emotional
  contagion through social networks.}}
\newblock {\em Proceedings of the National Academy of Sciences of the United
  States of America\/} {111}, 24 (June 2014), 8788--90.
\newblock
\showISSN{1091-6490}
\showDOI{%
\url{http://dx.doi.org/10.1073/pnas.1320040111}}


\bibitem{Landis77}
{J.~Richard Landis} {and} {Gary~G. Koch}. 1977.
\newblock \showarticletitle{The Measurement of Observer Agreement for
  Categorical Data}.
\newblock {\em Biometrics\/} {33}, 1 (1977).
\newblock


\bibitem{levallois:2013:SemEval-2013}
{Clement Levallois}. 2013.
\newblock \showarticletitle{Umigon: sentiment analysis for tweets based on
  terms lists and heuristics}. In {\em Second Joint Conference on Lexical and
  Computational Semantics (*SEM), Volume 2: Proceedings of the Seventh
  International Workshop on Semantic Evaluation (SemEval 2013)}. Association
  for Computational Linguistics, Atlanta, Georgia, USA, 414--417.
\newblock
\showURL{%
\url{http://www.aclweb.org/anthology/S13-2068}}


\bibitem{semantria2015}
{Lexalytics}. 2015.
\newblock {\em Sentiment Extraction - Measuring the Emotional Tone of Content}.
\newblock {T}echnical {R}eport. Lexalytics.
\newblock


\bibitem{liu2012sentiment}
{Bing Liu}. 2012.
\newblock \showarticletitle{Sentiment Analysis and Opinion Mining}.
\newblock {\em Synthesis Lectures on Human Language Technologies\/} {5}, 1 (May
  2012), 1--167.
\newblock
\showDOI{%
\url{http://dx.doi.org/10.2200/s00416ed1v01y201204hlt016}}


\bibitem{wordnet}
{George~A. Miller}. 1995.
\newblock \showarticletitle{WordNet: a lexical database for English}.
\newblock {\it Commun. ACM} {38}, 11 (1995), 39--41.
\newblock


\bibitem{mohammad:2012:STARSEM-SEMEVAL}
{Saif Mohammad}. 2012.
\newblock \showarticletitle{\#Emotional Tweets}. In {\em The First Joint
  Conference on Lexical and Computational Semantics - Volume 1: Proceedings of
  the main conference and the shared task, and Volume 2: Proceedings of the
  Sixth International Workshop on Semantic Evaluation (SemEval 2012)}.
  Association for Computational Linguistics, Montr\'eal, Canada, 246--255.
\newblock
\showURL{%
\url{http://www.aclweb.org/anthology/S12-1033}}


\bibitem{Mohammad-maryland2009}
{Saif Mohammad}, {Cody Dunne}, {and} {Bonnie Dorr}. 2009.
\newblock \showarticletitle{Generating High-coverage Semantic Orientation
  Lexicons from Overtly Marked Words and a Thesaurus}. In {\em Proceedings of
  the 2009 Conference on Empirical Methods in Natural Language Processing:
  Volume 2 - Volume 2} {\em (EMNLP '09)}. Association for Computational
  Linguistics, Stroudsburg, PA, USA, 599--608.
\newblock
\showISBNx{978-1-932432-62-6}
\showURL{%
\url{http://dl.acm.org/citation.cfm?id=1699571.1699591}}


\bibitem{journals/ci/MohammadT13}
{Saif Mohammad} {and} {Peter~D. Turney}. 2013.
\newblock \showarticletitle{Crowdsourcing a Word-Emotion Association Lexicon.}
\newblock {\em Computational Intelligence\/} {29}, 3 (2013), 436--465.
\newblock


\bibitem{MohammadKZ2013}
{Saif~M. Mohammad}, {Svetlana Kiritchenko}, {and} {Xiaodan Zhu}. 2013.
\newblock \showarticletitle{NRC-Canada: Building the State-of-the-Art in
  Sentiment Analysis of Tweets}. In {\em Proceedings of the seventh
  international workshop on Semantic Evaluation Exercises (SemEval-2013)}.
  Atlanta, Georgia, USA.
\newblock


\bibitem{semeval-2013task2}
{Preslav Nakov}, {Zornitsa Kozareva}, {Alan Ritter}, {Sara Rosenthal}, {Veselin
  Stoyanov}, {and} {Theresa Wilson}. 2013.
\newblock SemEval-2013 Task 2: Sentiment Analysis in Twitter.
\newblock   (2013).
\newblock


\bibitem{SaschaNarr2012}
{Sascha Narr}, {Michael Hülfenhaus}, {and} {Sahin Albayrak}. 2012.
\newblock \showarticletitle{Language-independent Twitter sentiment analysis}.
\newblock {\em Knowledge Discovery and Machine Learning (KDML)\/} (2012),
  12--14.
\newblock


\bibitem{nielsen2011new}
{Finn~{\AA}rup Nielsen}. 2011.
\newblock \showarticletitle{A new ANEW: Evaluation of a word list for sentiment
  analysis in microblogs}.
\newblock {\em arXiv preprint arXiv:1103.2903\/} (2011).
\newblock


\bibitem{conf/epia/OliveiraCA13}
{Nuno Oliveira}, {Paulo Cortez}, {and} {Nelson Areal}. 2013.
\newblock \showarticletitle{On the Predictability of Stock Market Behavior
  Using StockTwits Sentiment and Posting Volume.}
\newblock   {8154} (2013), 355--365.
\newblock
\showISBNx{978-3-642-40668-3}


\bibitem{Pang04asentimental}
{Bo Pang} {and} {Lillian Lee}. 2004.
\newblock \showarticletitle{A sentimental education: Sentiment analysis using
  subjectivity summarization based on minimum cuts}. In {\em In Proceedings of
  the ACL}. 271--278.
\newblock


\bibitem{Pang:2002:TUS:1118693.1118704}
{Bo Pang}, {Lillian Lee}, {and} {Shivakumar Vaithyanathan}. 2002.
\newblock \showarticletitle{Thumbs up?: sentiment classification using machine
  learning techniques}. In {\em ACL Conference on Empirical Methods in Natural
  Language Processing}. 79--86.
\newblock


\bibitem{Pappas_CICLING_2013}
{Nikolaos Pappas}, {Georgios Katsimpras}, {and} {Efstathios Stamatatos}. 2013.
\newblock \showarticletitle{Distinguishing the Popularity Between Topics: A
  System for Up-to-date Opinion Retrieval and Mining in the Web}. In {\em 14th
  International Conference on Intelligent Text Processing and Computational
  Linguistics}.
\newblock


\bibitem{pappas2013sentiment}
{Nikolaos Pappas} {and} {Andrei Popescu-Belis}. 2013.
\newblock \showarticletitle{Sentiment analysis of user comments for one-class
  collaborative filtering over TED talks}. In {\em Proceedings of the 36th
  international ACM SIGIR conference on Research and development in information
  retrieval}. ACM, 773--776.
\newblock


\bibitem{citeulike:8791184}
{R. Plutchik}. 1980.
\newblock {\em {A general psychoevolutionary theory of emotion}}.
\newblock Academic press, New York, 3--33.
\newblock


\bibitem{reis2015@icwsm}
{Julio Reis}, {Fabricio Benevenuto}, {Pedro Vaz~de Melo}, {Raquel Prates},
  {Haewoon Kwak}, {and} {Jisun An}. 2015.
\newblock \showarticletitle{Breaking the News: First Impressions Matter on
  Online News}. In {\em Proceedings of the 9th International AAAI Conference on
  Web-Blogs and Social Media (ICWSM)}.
\newblock


\bibitem{reis2014icwsm}
{Julio Reis}, {Pollyanna Goncalves}, {Pedro Vaz~de Melo}, {Raquel Prates},
  {and} {Fabricio Benevenuto}. 2014.
\newblock \showarticletitle{Magnet News: You Choose the Polarity of What you
  Read}. In {\em International AAAI Conference on Web-Blogs and Social Media}.
\newblock


\bibitem{sanders2011}
{Niek Sanders}. 2011.
\newblock Twitter Sentiment Corpus by Niek Sanders.
\newblock   (2011).
\newblock
\newblock
\shownote{\url{http://www.sananalytics.com/lab/twitter-sentiment/}.}


\bibitem{Snow:2008:CFG:1613715.1613751}
{Rion Snow}, {Brendan O'Connor}, {Daniel Jurafsky}, {and} {Andrew~Y. Ng}. 2008.
\newblock \showarticletitle{Cheap and Fast---but is It Good?: Evaluating
  Non-expert Annotations for Natural Language Tasks}. In {\em Proceedings of
  the Conference on Empirical Methods in Natural Language Processing} {\em
  (EMNLP '08)}.
\newblock


\bibitem{Socher-etal:2013}
{Richard Socher}, {Alex Perelygin}, {Jean Wu}, {Jason Chuang}, {Christopher~D.
  Manning}, {Andrew~Y. Ng}, {and} {Christopher Potts}. 2013.
\newblock \showarticletitle{Recursive Deep Models for Semantic Compositionality
  Over a Sentiment Treebank}. In {\em 2013 Conference on {E}mpirical {M}ethods
  in {N}atural {L}anguage {P}rocessing}. 1631--1642.
\newblock


\bibitem{Stone66}
{Philip~J. Stone}, {Dexter~C. Dunphy}, {Marshall~S. Smith}, {and} {Daniel~M.
  Ogilvie}. 1966.
\newblock {\em The General Inquirer: A Computer Approach to Content Analysis}.
\newblock MIT Press, USA.
\newblock


\bibitem{Strapparava:2007}
{Carlo Strapparava} {and} {Rada Mihalcea}. 2007.
\newblock \showarticletitle{SemEval-2007 Task 14: Affective Text}. In {\em
  Proceedings of the 4th International Workshop on Semantic Evaluations} {\em
  (SemEval '07)}. Association for Computational Linguistics, Stroudsburg, PA,
  USA, 70--74.
\newblock
\showURL{%
\url{http://dl.acm.org/citation.cfm?id=1621474.1621487}}


\bibitem{Taboada06a}
{Maite Taboada}, {Caroline Anthony}, {and} {Kimberly Voll}. 2006a.
\newblock \showarticletitle{Methods for Creating Semantic Orientation
  Dictionaries}. In {\em Conference on Language Resources and Evaluation
  (LREC)}. 427--432.
\newblock


\bibitem{Taboada:06a}
{Maite Taboada}, {Caroline Anthony}, {and} {Kimberly Voll}. 2006b.
\newblock \showarticletitle{Methods for Creating Semantic Orientation
  Dictionaries}. In {\em Conference on Language Resources and Evaluation
  (LREC)}. 427--432.
\newblock


\bibitem{Taboada:2011:LMS:2000517.2000518}
{Maite Taboada}, {Julian Brooke}, {Milan Tofiloski}, {Kimberly Voll}, {and}
  {Manfred Stede}. 2011.
\newblock \showarticletitle{Lexicon-based Methods for Sentiment Analysis}.
\newblock {\em Comput. Linguist.\/} {37}, 2 (June 2011), 267--307.
\newblock
\showISSN{0891-2017}


\bibitem{Tamersoy:2015}
{Acar Tamersoy}, {Munmun De~Choudhury}, {and} {Duen~Horng Chau}. 2015.
\newblock \showarticletitle{Characterizing Smoking and Drinking Abstinence from
  Social Media}. In {\em Proceedings of the 26th ACM Conference on Hypertext
  and Social Media (HT)}.
\newblock


\bibitem{TangWYZLQ14}
{Duyu Tang}, {Furu Wei}, {Nan Yang}, {Ming Zhou}, {Ting Liu}, {and} {Bing Qin}.
  2014.
\newblock \showarticletitle{Learning Sentiment-Specific Word Embedding for
  Twitter Sentiment Classification}. In {\em Proceedings of the 52nd Annual
  Meeting of the Association for Computational Linguistics, {ACL} 2014, June
  22-27, 2014, Baltimore, MD, USA, Volume 1: Long Papers}. 1555--1565.
\newblock


\bibitem{liwc}
{Yla~R. Tausczik} {and} {James~W. Pennebaker}. 2010.
\newblock \showarticletitle{The Psychological Meaning of Words: LIWC and
  Computerized Text Analysis Methods}.
\newblock {\em Journal of Language and Social Psychology\/} {29}, 1 (2010),
  24--54.
\newblock


\bibitem{sentistrength1}
{Mike Thelwall}. 2013.
\newblock Heart and soul: Sentiment strength detection in the social web with
  SentiStrength.
\newblock   (2013).
\newblock
\newblock
\shownote{\url{http://sentistrength.wlv.ac.uk/documentation/SentiStrengthChapter.pdf}.}


\bibitem{Tsytsarau:2012:SMS:2198180.2198208}
{Mikalai Tsytsarau} {and} {Themis Palpanas}. 2012.
\newblock \showarticletitle{Survey on Mining Subjective Data on the Web}.
\newblock {\em Data Min. Knowl. Discov.\/} {24}, 3 (May 2012), 478--514.
\newblock
\showISSN{1384-5810}
\showDOI{%
\url{http://dx.doi.org/10.1007/s10618-011-0238-6}}


\bibitem{Tumasjan}
{Andranik Tumasjan}, {Timm~O. Sprenger}, {Philipp~G. Sandner}, {and}
  {Isabell~M. Welpe}. 2010.
\newblock \showarticletitle{{Predicting Elections with Twitter: What 140
  Characters Reveal about Political Sentiment}}. In {\em International AAAI
  Conference on Weblogs and Social Media (ICWSM)}.
\newblock


\bibitem{Valitutti04wordnet-affect:an}
{Ro Valitutti}. 2004.
\newblock \showarticletitle{WordNet-Affect: an Affective Extension of WordNet}.
  In {\em In Proceedings of the 4th International Conference on Language
  Resources and Evaluation}. 1083--1086.
\newblock


\bibitem{sasa}
{Hao Wang}, {Dogan Can}, {Abe Kazemzadeh}, {Fran\c{c}ois Bar}, {and} {Shrikanth
  Narayanan}. 2012.
\newblock \showarticletitle{A system for real-time Twitter sentiment analysis
  of 2012 U.S. presidential election cycle}. In {\em ACL System
  Demonstrations}. 115--120.
\newblock


\bibitem{Watson}
{D. Watson} {and} {L. Clark}. 1985.
\newblock \showarticletitle{Development and validation of brief measures of
  positive and negative affect: the PANAS scales}.
\newblock {\em Journal of Personality and Social Psychology\/} {54}, 1 (1985),
  1063--1070.
\newblock


\bibitem{Wiebeetal05}
{Janyce Wiebe}, {Theresa Wilson}, {and} {Claire Cardie}. 2005.
\newblock \showarticletitle{Annotating Expressions of Opinions and Emotions in
  Language}.
\newblock {\em Language Resources and Evaluation\/} {1}, 2 (2005), 0.
\newblock
\showURL{%
\url{http://www.cs.pitt.edu/\~}}


\bibitem{OpinionFinder}
{Theresa Wilson}, {Paul Hoffmann}, {Swapna Somasundaran}, {Jason Kessler},
  {Janyce Wiebe}, {Yejin Choi}, {Claire Cardie}, {Ellen Riloff}, {and}
  {Siddharth Patwardhan}. 2005a.
\newblock \showarticletitle{OpinionFinder: a system for subjectivity analysis}.
  In {\em HLT/EMNLP on Interactive Demonstrations}. 34--35.
\newblock


\bibitem{wilson-wiebe-hoffmann:2005:HLTEMNLP}
{Theresa Wilson}, {Janyce Wiebe}, {and} {Paul Hoffmann}. 2005b.
\newblock \showarticletitle{Recognizing Contextual Polarity in Phrase-Level
  Sentiment Analysis}. In {\em ACL Conference on Empirical Methods in Natural
  Language Processing}. 347--354.
\newblock


\bibitem{Wolpert97nofree}
{David~H. Wolpert} {and} {William~G. Macready}. 1997.
\newblock \showarticletitle{No free lunch theorems for optimization}.
\newblock {\em IEEE Transactions on Evlutionary Computation\/} {1}, 1 (1997),
  67--82.
\newblock


\end{thebibliography}

\end{document}